\documentclass[letterpaper]{article}
\usepackage{uai2020}
\usepackage[margin=1in]{geometry}

\usepackage{times}

\usepackage[round]{natbib}

\usepackage{amsmath}
\usepackage{amssymb}
\usepackage{amsthm}
\usepackage{verbatim}

\newtheorem{proposition}{Proposition}

\usepackage[utf8]{inputenc} 
\usepackage[T1]{fontenc}    
\usepackage{hyperref}       
\usepackage{url}            
\usepackage{booktabs}       
\usepackage{amsfonts}       
\usepackage{nicefrac}       
\usepackage{microtype}      
\usepackage{algorithmic}
\usepackage{algorithm}

\usepackage{graphicx}
\usepackage{caption}
\usepackage{subcaption}
\usepackage{float}
\usepackage{multirow}

\usepackage{tikz}
\usetikzlibrary{arrows,positioning,fit,shapes,calc}

\newcommand{\E}{\mathrm{E}}
\DeclareMathOperator{\EX}{\E}
\DeclareMathOperator{\ND}{\mathcal{N}}

\title{Variance Constrained Autoencoding}

\author{ {\bf D. T. Braithwaite\thanks{\hspace{0.15cm} daniel.braithwaite@ecs.vuw.ac.nz},\hspace{0.25cm} M. O'Connor,\hspace{0.25cm}  W. B. Kleijn} \\
School of Engineering and Computer Science,\\
Victoria University of Wellington,\\
New Zealand \\
}

\begin{document}

\maketitle

\tikzstyle{box} = [rectangle, draw=black]
\tikzstyle{operation} = [diamond, draw=black]
\tikzstyle{arrow} = [thick,->,>=stealth]

\begin{abstract}
Recent state-of-the-art autoencoder based generative models have an encoder-decoder structure and learn a latent representation with a pre-defined distribution that can be sampled from. Implementing the encoder networks of these models in a stochastic manner provides a natural and common approach to avoid overfitting and enforce a smooth decoder function. However, we show that for stochastic encoders, simultaneously attempting to enforce a distribution constraint and minimising an output distortion leads to a reduction in generative and reconstruction quality. In addition, attempting to enforce a latent distribution constraint is not reasonable when performing disentanglement. Hence, we propose the variance-constrained autoencoder (VCAE), which only enforces a variance constraint on the latent distribution. Our experiments show that VCAE improves upon Wasserstein Autoencoder and the Variational Autoencoder in both reconstruction and generative quality on MNIST and CelebA. Moreover, we show that VCAE equipped with a total correlation penalty term performs equivalently to FactorVAE at learning disentangled representations on 3D-Shapes while being a more principled approach.
\end{abstract}

\section{Introduction}
\label{sec:introduction}
A common generative model is the variational autoencoder (VAE) \citep{kingma2013auto, rezende2014stochastic}. In recent papers, VAE was shown not to learn meaningful latent representations, i.e., the latent representation $z$ becomes statistically independent of the data $x$, if a sufficiently powerful probabilistic decoder is used \citep{bowman2015generating, chen2016variational, higgins2016beta, alemifixing}. A focus of many recent works has been to develop generative models based on VAE that learn meaningful latent representations \citep{tolstikhin2017wasserstein, braithwaite2018bounded, zhao2018infovae, alemifixing, razavi2019preventing}. Of these systems, the state-of-the-art Wasserstein Autoencoder (WAE) \citep{tolstikhin2017wasserstein} and Bounded Information Rate Variational Autoencoder (BIR-VAE) \citep{braithwaite2018bounded}, are conceptually relatively straightforward as they do not involve the explicit optimisation of information-theoretical measures. Consider an encoder $Q_{Z|X;\phi}$ and a decoder $P_{X|Z;\theta}$, with parameters $\phi$ and $\theta$ respectively, and let $X \sim P_D$ be the data distribution. WAE and BIR-VAE then minimise a mean output error $c$ and, additionally, attempt to drive the aggregate posterior distribution $q_\phi(z) = \int q_\phi(z|x) p_X(x) dx$ to a pre-defined prior $p(z)$, typically an isotropic Gaussian.

WAE authors advocated the use of deterministic encoders, where the variance of $Q_{Z|X;\phi}$ is zero. However, \citep{braithwaite2018bounded} show that using a fixed amount of encoder stochasticity (e.g., additive white gaussian noise) in the latent layer during training can be used to prevent overfitting in situations where limited data is available. Additionally, in section \ref{sec:drawbacks_of_cwae}, we discuss how using stochastic encoders during training results in the data-domain reconstruction cost $c$ encouraging neighbourhoods in the data domain to remain connected in the latent space. In contrast, the natural continuity of the encoder network is the only reason for such preservation of neighbourhood connectivity when the encoder is deterministic. This argument suggests that latent space noise is beneficial for generative modelling tasks, allowing for better sampling. Stochastic encoders, however, lead to some interesting challenges.

Consider an extension of WAE that uses stochastic encoders implemented by fixed variance additive noise in the latent layer. We view the stochastic WAE as transmitting latent codes, given by the function $\mu_\phi(x)$, through a noisy communication channel. The output of the channel, has the form $z = \mu_\phi(x) + \epsilon$, where $\epsilon \sim P_\epsilon$ is a user-defined distribution. The learned latent representation is jointly optimised to encode the most important information in the data  (\textit{source coding} \citep{cover2012elements}) and be robust to errors introduced by the noisy communication channel (\textit{channel coding} \citep{cover2012elements}), so as to minimise the distortion at the output.

WAE attempts to enforce a pre-specified shape on the aggregate posterior, $Q_{Z;\phi}$, by minimising the divergence $D[Q_{Z;\phi}||P_Z]$ \citep{tolstikhin2017wasserstein}. In the case of stochastic encoders, the learned aggregate posterior is a compromise between $P_Z$ and the distribution corresponding to the optimal code that minimises the expected distortion. This attempt to enforce the desired prior distribution, $P_Z$, prevents the optimal joint source-channel code from being learned, negatively effecting reconstruction performance. This behaviour also affects generative performance because $Q_{Z;\phi} \not= P_Z$, and, therefore, an incorrect distribution is assumed when sampling from the generative model. In addition, the regularisation of the aggregate posterior to match $P_Z$ is contrary to the objectives of disentanglement. In general, the true generative latent features for a dataset are not necessarily Gaussian. In section \ref{sec:experiments}, we observe that when disentangled features are learnt on 3D-Shapes \citep{3dshapes18}, they do not have a Gaussian distribution. Despite this, many state-of-the-art disentanglement methods \citep{higgins2017beta, chen2018isolating, kim2018disentangling} regularise the aggregate posterior to have a Gaussian distribution.

\textbf{Our Contributions:}
\begin{itemize}
	\item In section \ref{sec:drawbacks_of_cwae}, we demonstrate theoretically that in the case of the stochastic WAE, the compromise between the user-defined distribution $P_Z$ and the optimal code that minimises the expected distortion causes sub-optimal reconstruction and generative performance.
	\item In section \ref{sec:vcae}, we propose the Variance Constrained Autoencoder (VCAE) which applies only a variance constraint to the aggregate posterior rather than additionally constraining the shape (like WAE). Then, in section \ref{sec:experiments}, we demonstrate that VCAE outperforms the stochastic WAE (a state-of-the-art method) and VAE in terms of reconstruction quality and generative modelling on MNIST and CelebA.
	\item In section \ref{sec:experiments}, we show that VCAE equipped with a total correlation penalty (TC-VCAE) has equivalent performance to FactorVAE for the task of disentanglement on 3D-Shapes, while also a more principled disentanglement approach than FactorVAE.
\end{itemize}
\section{Constrained Wasserstein Autoencoder}
\label{sec:cwae}
In this section, we introduce two state-of-the-art latent variable generative models, the Wasserstein Autoencoder (WAE) \citep{tolstikhin2017wasserstein} and Bounded Information Rate Variational Autoencoder (BIR-VAE) \citep{braithwaite2018bounded}. Despite WAE and BIR-VAE being derived from different perspectives, in practice, they are equivalent. This relationship leads us to the constrained Wasserstein Autoencoder (cWAE), an information-rate limited WAE. 

Throughout the remainder of this paper, we denote random variables by capital letters, e.g., $X$, and their realisations as lower-case letters, e.g., $x$. Probability density functions $P(X=x) = p_X(x)$ are abbreviated to $p(x)$, and probability distributions are denoted as $P_X$. We will primarily deal with data and latent variables $x \in \mathbb{R}^{x_{dim}}$ and $z\in \mathbb{R}^{z_{dim}}$, respectively, where generally $x_{dim}>>z_{dim}$.

Both WAE and BIR-VAE have the same setup, with two stochastic mappings, $Q_{Z|X;\phi}$ and $P_{X|Z;\theta}$ implemented by neural networks with parameters $\phi$ and $\theta$ respectively. $Q_{Z|X;\phi}$ and $P_{X|Z;\theta}$ are referred to as the encoder and decoder, respectively. The aggregated posterior distribution is $Q_{Z;\phi} = \EX_{X \sim P_{X;D}}[Q_{Z|X;\phi}]$, where $P_{X;D}$ is the distribution of $X$ defined by the data. The aggregated generative distribution is $P_{X;\theta} = \EX_{Z \sim Q_{Z;\phi}}[P_{X|Z;\theta}]$. Lastly, let $P_Z$ be a user-selected distribution.

WAE optimises this aforementioned model by minimising the Wasserstein distance between $P_{X;D}$ and $P_{X;\theta}$. In general, the Wasserstein distance is not easily computable. However, when the decoder is implemented by a deterministic function, denoted $\mu_\theta(z)$, the Wasserstein distance can be written as \citep{bousquet2017optimal}:
\begin{equation}
	\underset{Q_{Z|X;\phi}:\ Q_{Z;\phi} = P_Z}{\inf} \EX_{X \sim P_D}\EX_{Z \sim Q_{Z|X;\phi}} [c(X, \mu_\theta(Z))],
	\label{equ:wae_constrained_opt}
\end{equation}
where $c(\cdot, \cdot)$ can be any \textit{distance metric}, and $Q_{Z;\phi}$ is constrained to match the user-defined distribution $P_Z$ (e.g., $\ND(0, I_{z_{dim}})$). WAE takes advantage of this convenient form of the Wasserstein distance.

The distribution constraint in \eqref{equ:wae_constrained_opt}, that $Q_{Z;\phi} = P_Z$, cannot be enforced directly, and must be relaxed using a penalty function. Hence, \eqref{equ:wae_constrained_opt} is written as the following unconstrained optimisation problem for WAE \citep{tolstikhin2017wasserstein} objective:
\begin{equation}
\label{equ:wae_objective_optimisable}
\begin{aligned}
&\underset{\phi, \theta}{\text{minimise}}&&\EX_{X \sim P_D} \EX_{Z \sim Q_{Z|X;\phi}}[c(X, \mu_\theta(Z)] \\&&&+ \lambda D_z[Q_{Z;\phi} || P_Z],
    \end{aligned}
\end{equation}
where $D_z$ is a divergence and $\lambda$ (a hyper-parameter) controls the trade-off between minimising the expected distortion and attempting to enforce the constraint. 

We now discuss two motivations for stochastic encoders. Firstly, it was shown that fixed-variance stochastic encoders could be used to prevent overfitting when limited data is available \citep{braithwaite2018bounded}. Secondly, stochastic encoders explicitly prioritise the local structure in the data domain to be expressed in the latent structure, improving generalisation. We expand on this in section \ref{sec:drawbacks_of_cwae}.

One approach for implementing the stochastic encoders is using fixed additive noise \citep{braithwaite2018bounded}. In this case, for a given $x$, $Z \sim Q_{Z|x;\phi}$ has the form $z = \mu_\phi(z) + \epsilon$ where $\epsilon \sim P_{\epsilon}$ and $P_{\epsilon}$ is a user-defined distribution. WAEs with stochastic encoders can also be implemented \citep{rubenstein2018learning} using the reparameterisation trick \citep{kingma2013auto, rezende2014stochastic}. In this case, $Q_{Z|X;\phi}$ is a diagonal Gaussian and both the mean and variance of $Q_{Z|X;\phi}$ are a function of $X$. A disadvantage with the latter approach is that in practice, the variance of $Q_{Z|X;\phi}$ can decay to 0 \citep{rubenstein2018learning}, removing any noise in the latent layer. We chose to implement the stochastic encoders using the method of \citep{braithwaite2018bounded}, primarily because fixing the variance of the noise means it cannot decay to 0. Additionally, the method of \citep{braithwaite2018bounded} is simpler than the of approach \citep{rubenstein2018wasserstein} and, as discussed below, allows the mutual information between $X$ and $Z$ to be explicitly controlled.

A constrained optimisation problem for WAE with Gaussian stochastic encoders can now be formulated. Let $P_\epsilon = \ND(0, \sigma_\epsilon^2 \cdot I_{z_{dim}})$, where $\sigma_\epsilon^2$ is a hyper-parameter. Additionally, let $P_Z = \ND(0, I_{z_{dim}})$, as is commonly done. Then, an extended version of WAEs objective \eqref{equ:wae_objective_optimisable} that includes stochastic encoders can be written as:
\begin{equation}
\label{equ:c_wae_objective_optimisable}
\begin{aligned}
&\underset{\phi, \theta}{\text{minimise}}&&\EX_{X \sim P_D} \EX_{Z \sim Q_{Z|X;\phi}}[c(X, \mu_\theta(Z)] \\&&&+ \lambda D_z[Q_{Z;\phi} \| \ND(0, I_{z_{dim}})] \\
&\text{subject to}&& \EX_{Q_{Z|X;\phi}}[ (Z - \EX_{Q_{Z|X;\phi}}[Z])^2 ] =   \sigma^2_\epsilon I_{z_{dim}}.
    \end{aligned}
\end{equation}

We now introduce concepts relating to source and channel coding. Let $\mathcal{X}$ be a set of data-points, then the function $C: \mathcal{X} \rightarrow \Sigma$ is a \textit{code}, where $\Sigma$ is an alphabet of codewords. For $x \in \mathcal{X}$, $C(x)$ is the codeword associated with $x$. A \textit{communication channel} is defined as $p(z|C(x))$, where $z$ is the output of the channel. The objective of channel coding is to minimise the overall distortion (e.g., L1 or L2 error) between the input and reconstructed data points after the channel. The \textit{channel capacity}, M, is the theoretical maximum amount of information that can be transmitted through the channel. Only in the case of infinite-delay does the optimal code achieve the maximum channel capacity. In the cases of finite-delay, the optimal code will not necessarily reach this bound. To compute $M$, first we construct the joint distribution $P_{X, C(X)} = P_{X|C(X)} \cdot P_{C(X)}$, then $M$ is given by:
\begin{equation}
	M = \sup_{P_{C(X)}} I(X; C(X)),
	\label{equ:general_information_rate}
\end{equation}
under a constraint on the power of transmission, i.e., the summed variance of each dimension of $Z = C(X)$ is constrained to be $v$, a hyper-parameter. Since $I_\phi(Z;C(X)) = H(C(X)) - H(Z|C(X))$, for a fixed channel $P_\epsilon$, $M$ corresponds to choosing $Q_{Z;\phi}$ with maximum entropy. Therefore, for infinite-delay, the optimal distribution of codewords is a symmetric Gaussian.

A stochastic WAE can be interpreted as transmitting a code, given by $\mu_\phi(x)$, through a noisy AWGN channel. The channel is defined by $z = \mu_\phi(x) + \epsilon$, where $\epsilon \sim \ND(0, \sigma_\epsilon^2)$ and $\sigma_\epsilon^2$ is a hyper-parameter. Hence, for a fixed $Q_\phi(Z)$, increasing $\sigma_\epsilon^2$ decreases the theoretical channel capacity, limiting the number of bits that can be transmitted by the latent layer. We denote this model the constrained Wasserstein Autoencoder (cWAE), as it is information rate limited. cWAE is equivalent to the Bounded Information Rate Variational Autoencoder. 

cWAE attempts to enforce a Gaussian distribution on the output of the encoder, which in the infinite-delay case would correspond to the optimal specification of the latent code. However, this situation has finite-delay, and as a consequence, this attempt to specify the shape of the aggregate posterior causes a reduction in reconstruction and generative performance. We discuss the disadvantages of cWAE in the following section.
\section{Drawbacks of cWAE}
\label{sec:drawbacks_of_cwae}
In this section, we introduce and discuss the drawbacks of cWAE. Primarily, this section looks at how the specification of a desired latent distribution causes a higher expected distortion at the output.

We first build on concepts introduced in the previous section with the following definitions. A \textit{source coder} compresses the input dataset, removing redundancy to express data points in as few bits as possible. On the other hand, a \textit{channel coder} introduces redundancies to make the codewords robust to transmission over a noisy communication channel. Lastly, a \textit{joint source-channel coder} performs both source and channel coding simultaneously. The \textit{separation theorem} \citep{shannon1948mathematical} proves that source and channel codes can be optimised independently. However, this theorem relies on infinitely long codes, something that does not hold in practice. Consequently, using a joint source-channel coder can lead to lower expected distortion compared with performing source and channel coding individually.

We now return to the interpretation of cWAE as transmitting latent codes across a communication channel and minimising the expected distortion. In this interpretation, the encoder function $\mu_\phi(x)$ represents the code and $z = \mu_\phi(x) + \epsilon$ represents its noisy transmission through the channel, where $\epsilon \sim P_\epsilon$ is a zero-mean distribution defined by the user. Hence, cWAE is a joint source-channel coder. Learning binary joint source-channel codes has previously been explored \citep{choi2018necst}.

Typically, cWAE attempts to enforce a Gaussian distribution on its aggregate posterior, maximising the amount of information being transmitted across the channel for a given restriction on the transmission power. In the case of infinite delay, this would correspond to the optimal distribution of codewords. However, the situation we are considering is not infinite-delay, but rather finite-delay. For transition over an AWGN communication channel with finite-delay, a Gaussian distribution at the output of the encoder is optimal (results in minimum mean square error) only if the input data are Gaussian. In general, the optimal distribution of codewords is dependent on the data distribution \citep{akyol2010optimal}. In the following arguments, we first decompose the mean reconstruction cost used as the objective function for cWAE, revealing why stochastic encoders cause preservation of data domain connectivity in the learned representation. These observations then lead to the result that the latent representation depends on the data distribution. Hence, the conclusion that attempting to enforce a distribution on the aggregate posterior negatively affects reconstruction and generative performance.

\begin{proposition}
	Let $\mu_\phi: X \rightarrow Z$ and $\mu_\theta: Z \rightarrow X$ be  differentiable functions, and let $\epsilon \sim P_\epsilon$ be a zero-mean distribution with variance $\sigma_\epsilon^2$. The encoding operation is defined as $Z = \mu_\phi(X) + \epsilon$, and the decoding operation is $\hat{X} = \mu_\theta(Z)$. Then, for sufficiently small $\sigma_\epsilon^2$:
	\begin{multline}
		\E_{\epsilon} [ \| x - \hat{x} \|^2_2] = \|x - \mu_\theta(\mu_\phi(x)) \|_2^2 \\+ \sigma_\epsilon^2 \| J_{\mu_\theta}(\mu_\phi(x)) \|^2_F,
		\label{equ:analytic_approx_of_vcae_obj}
	\end{multline}
	where $J_{\mu_\theta}$ is the Jacobian of the decoder function $\mu_\theta$.
	\label{lem:jacobian_objective}
\end{proposition}

Proposition \ref{lem:jacobian_objective} (proof in section \ref{sec:prop_jacobian_objective_proof} of the appendix) shows that minimising the mean square reconstruction cost under the imposed conditions results in an objective regularised by the Jacobian of the decoder. A similar idea has been previously explored twice. The first case is the de-noising autoencoder \citep{vincent2008extracting}, where adding noise in the input domain results in an objective similar to \eqref{equ:analytic_approx_of_vcae_obj}, but minimises $\| J_{\mu_\theta \circ \mu_\phi}(X) \|^2_F$, instead of $\| J_{\mu_\theta}(\mu_\phi(x)) \|^2_F$ \citep{bishop1995training}. The second case is the Contractive Autoencoder (CAE) \citep{rifai2011contractive}, which explicitly adds the Frobenius norm of the decoder's Jacobian to the standard autoencoder objective. The authors of CAE find this regularisation results in a learned representation that better represents the data and is robust to perturbations in the input domain.

The squared Frobenius norm of a matrix is equal to the sum of the squared singular values. Additionally, the Lipschitz norm of a matrix is given by its largest singular value. Therefore,  for a matrix A, we have the following:
\begin{align*}
	\|A\|_F^2 = \sum s_i(A) \geq \max s_i(A) = \|A\|_{Lip},
\end{align*}
where $s_i$ is the $i$th singular value. Consequently, $\| J_{\mu_\theta}(\mu_\phi(x)) \|^2_F$ is an upper bound on the Lipschitz value of the local transform defined by the Jacobian of $\mu_\theta$. This analysis affords the interpretation of the objective as regularising the local Lipschitz value of $\mu_\theta(z)$ for the \textit{neighbourhood} of each $\mu_\phi(x)$. The expressive power of $\mu_\theta(z)$ is restricted by this regularisation. This argument also demonstrates why VCAE preserves data-domain connectivity in the latent representation: optimising the mean squared error resulting from a small noise addition $\epsilon$ in Z means that the system attempts to make a neighbourhood of any realisation of $Z$ correspond to a neighbourhood of $\hat{X} = \mu_\phi^{-1}(Z)$. The final result is that nearby points in Z are nearby points in X and that the mapping is, therefore, smooth. If the latent distribution is left entirely unconstrained, then its variance will grow without bound. Hence, at minimum, a constraint on the variance is required. If both $P_D$ and the variance of $Z$ are specified, then an optimisation problem for $Q_{Z;\phi}$ is obtained. 

Preservation of data-domain connectivity in the latent representation necessarily means that the structure of the aggregate posterior depends on the data distribution. Therefore, we see that when enforcing a pre-specified distribution, $P_Z$, on the aggregate posterior, $Q_{Z;\phi}$, the resultant learned shape will be a compromise between $P_Z$ and what is optimal for minimising the mean reconstruction cost. Hence, causing degradation in both reconstructive and generative performance. Reconstruction performance is affected because the mean reconstruction cost is not minimised. On the other hand, generative performance is affected because the latent distribution is not equal to what was enforced. Thus, when sampling from the model, an incorrect latent distribution is assumed.
\section{Variance Constrained Autoencoder}
\label{sec:vcae}

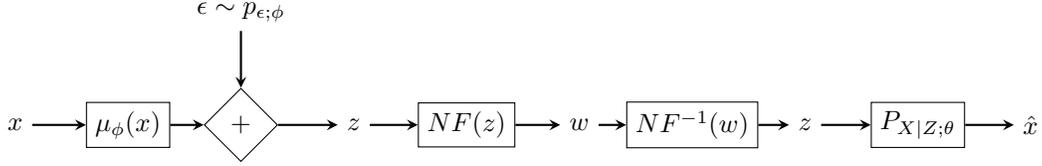
\begin{figure*}
\centering
	\begin{tikzpicture}[node distance=1.5cm]
		\node (data) [] {$x$};
		\node (encnet) [box, right of=data] {$\mu_\phi(x)$};
		\draw [arrow] (data) -- (encnet);
		\node (noise) [operation, right of=encnet] {$+$};
		\node (encnoisedist) [above of=noise] {$\epsilon \sim p_{\epsilon;\phi}$};
		\draw [arrow] (encnet) -- (noise);
		\draw [arrow] (encnoisedist) -- (noise);
		\node (lf) [right of=noise] {$z$};
		\draw [arrow] (noise) -- (lf);
		\node (nf) [box, right of=lf] {$NF(z)$};
		\draw [arrow] (lf) -- (nf);
		\node (w) [right of=nf] {$w$};
		\draw [arrow] (nf) -- (w);
		\node (nfinv) [box, right of=w] {$NF^{-1}(w)$};
		\draw [arrow] (w) -- (nfinv);
		\node (lfdec) [right of=nfinv] {$z$};
		\draw [arrow] (nfinv) -- (lfdec);
		\node (decnet) [box, right of=lfdec] {$P_{X|Z;\theta}$};
		\draw [arrow] (lfdec) -- (decnet);
		\node (recon) [right of=decnet] {$\hat{x}$};
		\draw [arrow] (decnet) -- (recon);
	\end{tikzpicture}
	
	\caption{VCAE Architecture. In the diagram the normalising flows are given by $NF(z) = f_w(\dots f_1(z)\dots )$.}
	\label{fig:vcae_archetchure}
\end{figure*}

In this section, we introduce the Variance Constrained Autoencoder (VCAE), a generative model with the same structure as cWAE, but which applies a variance constraint to the aggregate posterior $Q_{Z;\phi}$, rather than constraining its shape. The motivation for this change in constraints is the argument presented in section \ref{sec:drawbacks_of_cwae}, where we discuss issues with enforcing a shape on the latent distribution when using stochastic encoders.

The Variance Constrained Autoencoder (VCAE) is made up of two probabilistic mappings, $Q_{Z|X;\phi}$ and $P_{X|Z;\theta}$, the encoder and decoder, respectively. $Q_{Z|X;\phi}$ and $P_{X|Z;\theta}$ are implemented by neural networks with parameters $\phi$ and $\theta$, respectively. The probabilistic encoder is implemented by adding noise to a deterministic mapping: $z \sim Q_{Z|x;\phi}$ has the form $z = \mu_\phi(x) + \epsilon$, where $\epsilon \sim P_{\epsilon}$ is a user-defined distribution. It is common to use the mean of the decoder as output \citep{braithwaite2018bounded, tolstikhin2017wasserstein} denoted $\hat{x} = \mu_\theta(z)$, and we do so here as well. The aggregate posterior and generative distributions are defined as before.

The principle of VCAE is to maximise the likelihood of the data while constraining the variance of the aggregate posterior. This is in contrast to WAE (and BIR-VAE), where $Q_{Z;\phi}$ is regularised to be a pre-specified distribution. We write VCAE's objective as:
\begin{equation}
\label{equ:vcae-constrained-objective}
\begin{aligned}
&\underset{\phi, \theta}{\text{maximize}}&&\EX_{X \sim P_D} \EX_{Q_{Z|X;\phi}}[\log p_{X|Z;\theta}(X|Z)]\\
&\text{subject to}&&\EX_{Z \sim Q_{Z; \phi}}[||Z - \EX_{Z \sim Q_{Z; \phi}}[Z]||_2^2] = v,
 \end{aligned}
\end{equation}
$P_D$ is the data distribution and $v$ a hyper-parameter specifying the desired total variance. We relax the constraint in \eqref{equ:vcae-constrained-objective} using a penalty function, giving:
\begin{multline}
\label{equ:vcae-objective}
\underset{\phi, \theta}{\text{maximize}} \quad \EX_{X \sim P_D} \EX_{Z \sim Q_{Z|X;\phi}}[\log
  p_\theta(X|Z)] \\- \lambda \, |\EX_{Z \sim Q_{Z; \phi}}[||Z - \EX_{Z \sim Q_{Z; \phi}}[Z]||_2^2] - v|,
\end{multline}
an unconstrained optimisation problem where $\lambda$ is a hyper-parameter controlling the trade-off between maximising the likelihood and approximating the variance constraint. The variance penalty is computed per batch. 
 
We can similarly view VCAE as transmitting datapoint encodings over a noisy communication channel. The code is given by $\mu_\phi(x)$ and the channel is defined by the choice of the distribution $P_\epsilon$, where the output from the channel is $z = \mu_\phi(x) + \epsilon$. Therefore, like cWAE, this affords the interpretation of VCAE as a joint source-channel coder. However, in this case, only the variance of the aggregate posterior is constrained. This is in contrast to the cWAE which restricts the shape to be that of a pre-defined distribution $P_Z$. This change in constraints means that the learned latent distribution is no longer a compromise between the desired prior, $P_Z$, and the optimal distribution of the joint source-channel code that minimises the expected distortion. Consequently, for VCAE, a lower distortion can be achieved. 

For WAE, the aggregate posterior $Q_{Z;\phi}$ is regularised to be the user-defined distribution $P_Z$, which theoretically allows for easy sampling from the generative model. In the case of VCAE, the aggregate posterior $Q_{Z;\phi}$ is not known and therefore sampling from the trained generative model is not directly possible. To facilitate sampling, a chain of normalising flows transforms $Q_{Z;\phi}$ into a known distribution (selected by the user before training). Since normalising flows are invertible, this transform can be undone before the decoder. Consequently, this operation does not affect the training of the encoder or decoder and thus can be trained as a subsequent step. Moreover, if sampling is not required, then these normalising flows need not be trained. 

Let $P_W$ be a user-defined distribution (e.g. unit Gaussian), which will be transformed into our aggregate posterior $Q_{Z;\phi}$ using a chain of normalising flows. Let $f_1, ..., f_m$ be the set of invertible and continuous functions. Denote $w_n = f_n(w_{n-1})$, where $w_0 \sim P_W$, this induces the p.d.f $\hat{q}_\psi(w_m)$ (where $\psi$ represents the parameters of the $f_i$'s), which we can write as:
\begin{equation}
	\hat{q}_\psi(w_m) = p_w(w_0) \prod_{k=1}^{m} |\det \frac{\partial f_k}{\partial w_{k-1}}|^{-1}.
\end{equation}
We wish to optimise the functions $f_1, ..., f_m$ so that $\hat{Q}_{W_m; \psi} = Q_{Z; \phi}$, something that can be achieved by maximising the log-likelihood of samples mapped inversely though the normalising flows, given by:
\begin{equation}
\label{equ:norm_flow_objective}
\begin{aligned}
&\underset{\psi}{\text{maximize}}&&\EX_{Z \sim Q_{Z; \phi}} [\log
  p_{W}(f_1^{-1}(\cdot \cdot \cdot f_m^{-1}(Z) \cdot \cdot \cdot ))].
 \end{aligned}
\end{equation}
We choose normalising flows to implement this transformation because they are well defined and provide a convenient method for constructing an invertible transform from a known p.d.f $p(w)$ to the unknown p.d.f that describes our learned latent representation $q_{z;\phi}(z)$. However, it is important to note that the VCAE system is not restricted to normalising flows, and any method of approximating the aggregate posterior can be used. Figure \ref{fig:vcae_archetchure} is a diagram of the VCAE architecture and Algorithm \ref{alg:vcae_algorythm} (given in section \ref{sec:exp_setup} of the supplementary material) describes an implementation of the VCAE, where we assume $P_{X|Z;\theta}$ is a symmetric Gaussian permitting the use of the mean squared error (MSE) at the decoder output. 

While not immediately knowing $Q_{Z;\phi}$ is a limitation of VCAE, this issue is often also present for cWAE. In the case of cWAE, the aggregated posterior $Q_{Z;\phi}$ is regularised to be $P_Z$. However, after training, $Q_{Z;\phi}$ is a compromise between $P_Z$ and the distribution of latent vectors that is optimal with respect to minimising the expected reconstruction error. Therefore, the aggregate posterior is also not known in the case of cWAE. This result is demonstrated experimentally in section \ref{sec:aux_exp:latent_space_analysis} of the supplementary material, and by the discrepancy between the FID scores for WAE with the assumed prior $P_Z$ and approximated aggregate posterior (shown in table \ref{tab:mnist_celeba_results}).

Similarly to cWAE, \eqref{equ:general_information_rate} shows that there exists an upper bound on the mutual information between $X$ and $Z$ for VCAE. In the experimental section of this paper, we choose $P_\epsilon = \ND(0, \sigma_\epsilon^2 \cdot I_{z_{dim}})$ and $v = z_{dim}$, where $\sigma_\epsilon^2 \leq 1$ is user-defined. In this case, the upper bound on the information rate can be computed as: $I_{bits} = \frac{z_{dim}}{2}\log_2(\frac{1}{\sigma_\epsilon^2})$. For a fixed $v$, increasing $\sigma_\epsilon^2$ decreases the maximum information rate. On the other hand, decreasing $\sigma_\epsilon^2$ allows higher information throughput. However, setting $\sigma_\epsilon^2$ too low can allow overfitting \citep{braithwaite2018bounded}.

As previously mentioned, VCAE is a natural model for learning disentangled representations. This is because: 1) VCAE allows a flexible latent distribution; 2) Using stochastic encoders enforces a smooth latent representation, in which local neighbourhoods of points in the data domain are maintained in the latent space. To facilitate disentanglement, a penalty term can be added to VCAE's objective which enforces independence between $z_1, ..., z_{z_{dim}}$, the different latent features. The penalty term used will be a total correlation (TC) penalty term: $D_{KL}[Q_{Z;\phi}||\prod_i Q_{Z_i;\phi}]$, following \citep{kim2018disentangling, chen2018isolating}. To implement the TC penalty, we use the method of \citep{kim2018disentangling}. We denote VCAE equipped with the TC penalty as Total Correlation VCAE (TC-VCAE). The impossibility result of \citep{locatello2018challenging} states that disentanglement is, in general, impossible for factorised priors and that the reason many disentanglement methods work in practice is because of their implicit assumptions of the model. VCAE also enforces a factorised prior, but optimises for the conserving of data-domain connectivity in the latent space.

\section{Related Work}
\label{sec:related_work}
VAE + NF \citep{rezende2015variational} and VAE + IAF \citep{kingma2016improved} are two extensions on VAE which apply normalising flows to the distribution $Q_{Z|X;\phi}$ during training, to allow a more flexible latent distribution. VCAE relies on normalising flows to facilitate sampling from the trained generative model. However, these flows can be trained as a secondary process and are only necessary when sampling is required; this is in contrast to VAE + NF/IAF, where the flows are always needed and must be trained along with the rest of the system. 

Two recent disentanglement models, FactorVAE \citep{kim2018disentangling} and $\beta$-TCVAE \citep{chen2018isolating} extend VAE with a total correlation penalty term. These aforementioned methods have been shown to perform well on disentanglement tasks. However, both methods maximise the ELBO, which can be seen as actively working against the task of disentanglement because the KL-divergence term is at a minimum when $X$ and $Z$ are independent. WAE has been used for disentanglement by enforcing a factorised prior \citep{rubenstein2018learning}. In addition, these aforementioned methods all attempt to enforce a user-defined shape on the aggregate posterior. In section \ref{sec:vcae_disentanglemetn}, we show that this constraint does not align with the objectives of disentanglement. TC-VCAE is, therefore, a more principled disentanglement approach as it allows the shape of the aggregate posterior to vary.

\citep{ghosh2019variational} implement VAE's encoder deterministically instead applying L2 regularisation, spectral normalisation \citep{miyato2018spectral}, or gradient penalty \citep{gulrajani2017improved} to the decoder function. The proposed model was shown to outperform the VAE and WAE in terms of reconstruction and generative performance. However, unlike VCAE, the method proposed in \citep{ghosh2019variational} still enforces a prior distribution, and does not consider the associated information-theoretic disadvantages.

Generative Latent Flow (GLF) \citep{xiao2019generative} was also developed concurrently to VCAE. In GLF, a deterministic autoencoder is trained and regularised by a normalising flow mapping from the AE latent space to a unit Gaussian. While GLF is similar in structure to VCAE, \citep{xiao2019generative} do not address the importance of latent noise, the relationships to cWAE and information theory nor is a connection to disentanglement made.

\section{Experiments}
\label{sec:experiments}
In this section, we compare VCAE against VAE, VAE + IAF and WAE. A complete description of the experimental setup, including network architectures and hyper-parameters, is given in Appendix \ref{sec:exp_setup}. A description of how the models are compared fairly is also given in the appendix. To summarise, we select the VCAE variance constraint such that it has the same maximum encoding channel capacity as cWAE. Additionally, we select hyper-parameter settings for all models, which result in the respective constraints being sufficiently enforced. Sample implementations for VCAE are available.\footnote{See supplementary material for code.}

We first give two toy examples in section \ref{sec:toy_experiment}, demonstrating the efficacy of VCAE over cWAE. Then, in section \ref{sec:results_overfitting}, we experiment with preventing overfitting using stochastic encoders. In section \ref{sec:generative_modeling_exp}, we evaluate VCAE's generative modelling performance. Next, in section \ref{sec:vcae_disentanglemetn}, we equip VCAE with a total correlation penalty term (denoted TC-VCAE) and evaluate its disentanglement performance. An auxiliary experiment in section \ref{sec:aux_exp:latent_space_analysis} of the supplementary material investigates the structure of the latent space for VCAE and cWAE on MNIST, before and after applying the normalising flows.

\subsection{Toy Experiments}
\label{sec:toy_experiment}
\begin{figure*}
	\centering
	\begin{subfigure}[t]{0.15\textwidth} \includegraphics[width=\textwidth]{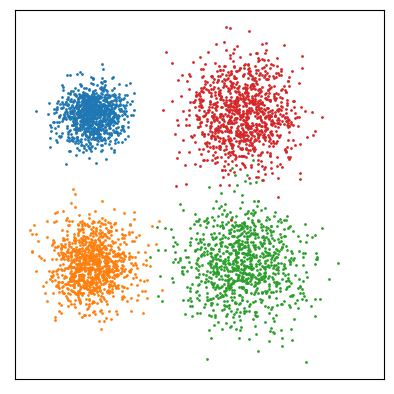} \caption{Data.} \label{fig:mog:toy_data} 
	\end{subfigure}
	\begin{subfigure}[t]{0.15\textwidth} \includegraphics[width=\textwidth]{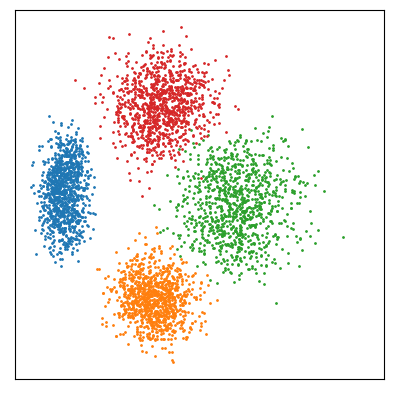} \caption{VCAE.} \label{fig:mog:vcae} 
	\end{subfigure}
	\begin{subfigure}[t]{0.15\textwidth} \includegraphics[width=\textwidth]{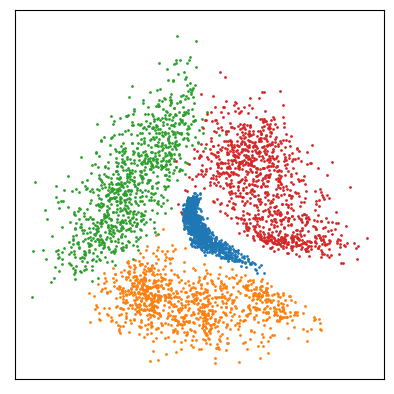} \caption{cWAE.} \label{fig:mog:cwae} 
	\end{subfigure}
	\hspace{0.05\textwidth}
	\begin{subfigure}[t]{0.15\textwidth} \includegraphics[width=\textwidth]{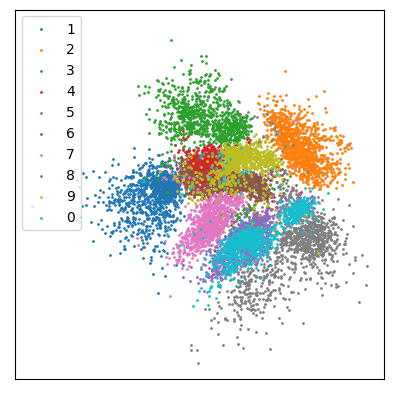} \caption{VCAE.} \label{fig:mnist:vcae} 
	\end{subfigure}
	\begin{subfigure}[t]{0.15\textwidth} \includegraphics[width=\textwidth]{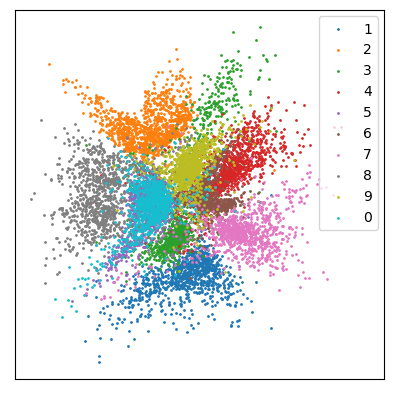} \caption{cWAE.} \label{fig:mnist:cwae} 
	\end{subfigure}
	
	\caption{Left three figures are the underlying MoG data and representations learned by VCAE and cWAE on a Mixture of Gaussians dataset. Right two figures are representations learned by VCAE and cWAE on MNIST.}
	\label{fig:toy_experiment}
\end{figure*}

In this section, we present two toy experiments that show specific situations where VCAE outperforms cWAE. First, we consider a dataset generated from a mixture of four two-dimensional Gaussians (MoG)(shown in figure \ref{fig:mog:toy_data}) and projected into a 100-dimensional space. Secondly, we look at models trained on MNIST with a two-dimensional latent space. These toy examples demonstrate the need for and utility of VCAE. In both cases, attempting to enforce a Gaussian distribution on $Q_{Z;\phi}$ causes a degradation in performance.

Figures \ref{fig:mog:vcae} \& \ref{fig:mog:cwae} show the representations learned by VCAE and cWAE respectively, when trained on this MoG dataset. cWAE enforces a Gaussian distribution, but this comes at the cost of performance, cWAE and VCAE achieve 0.13 and 0.067 MSE, respectively.

For the second experiment, we look at how these methods perform on the MNIST dataset, with two latent features to facilitate analysis. We find that a large regularisation parameter is required to ensure the desired distribution is enforced for cWAE. Moreover, this prior regularisation negatively affects performance. VCAE achieves training/testing reconstruction errors of 25.52/27.67, and cWAE achieves 30.09/31.50. Figures \ref{fig:mnist:vcae} \& \ref{fig:mnist:cwae} show the latent representations learned by VCAE and cWAE, respectively. Histograms of the features learned by VCAE and cWAE are provided in section \ref{sec:aux:toy_mnist_example} of the appendix. These histograms confirm that cWAE's distribution is close to a unit Gaussian, while VCAE's is not.

\subsection{Overfitting}
\label{sec:results_overfitting}
In this section, we trained VCAE, (c)WAE and VAE on the ReducedMNIST problem, a 600 element subset of the MNIST training data, to demonstrate that the use of stochastic encoders can prevent overfitting. For these experiments let dVCAE refer to VCAE with a deterministic encoder. The experimental setup for the following experiments is in section \ref{sec:setup:overfitting} of the supplementary material.

Table \ref{tab:overfitting_exp} show the results from these experiments, demonstrating that using stochastic encoders does indeed reduce the degree of overfitting as both VCAE and cWAE improve on WAE and dVCAE. Additionally, VCAE further reduces the amount of overfitting when compared to an equivalent (c)WAE. Lastly, we see the benefit of fixed rather than variable latent noise by observing that VCAE and cWAE improve upon VAE. In this case, VAE had a larger degree of overfitting because the variance in the latent dimension can be driven to zero, maximising performance on the training set but reducing generalisation.

\begin{table}[h!]
	\centering
	\caption{Training/testing errors for several models trained on the ReducedMNIST dataset. The noise distribution is given in brackets next to the model name.}
	\label{tab:overfitting_exp}
	\begin{tabular}[t]{ccc}
	Model & Train Err & Test Err\\
	\cline{1-3}
	VCAE ($\epsilon \sim \ND(0, 0.2)$) & 22.16 & 41.96 \\
	cWAE ($\epsilon \sim \ND(0, 0.2)$) & 18.06 & 44.56 \\
	\cline{1-3}
	VCAE ($\epsilon \sim \ND(0, 0.1)$) & 17.02 & 47.39 \\
	cWAE ($\epsilon \sim \ND(0, 0.1)$) & 15.19 & 47.39 \\
	\cline{1-3}
	dVCAE ($\epsilon = 0$) & 16.56 & 53.16 \\
	WAE ($\epsilon = 0$) & 14.79 & 58.50 \\
	\hline
	VAE & 10.45 & 53.32 \\
	\end{tabular}
\end{table}

\subsection{Generative Quality: MNIST \& CelebA}
\label{sec:generative_modeling_exp}
\begin{table*}[h!]
\centering
\caption{Experimental results show training/testing error and FID scores (smaller is better) for models trained on MNIST and CelebA. In the case of FID scores, the column labelled $P_{Z}$ refers to assuming the latent distribution is $\ND(0, I_{z_{dim}})$ and the column $\hat{Q}_{Z;\phi}$ refers to when the latent distribution is learned using normalising flows. FID scores are computed using 10000 testing set images and sampled images.}
\label{tab:mnist_celeba_results}

\begin{tabular}[t]{cccccccccccc}
 &\multicolumn{5}{c}{MNIST} & &\multicolumn{5}{c}{CelebA}\\
 \cline{2-6}\cline{8-12}
 &\multicolumn{2}{c}{Reconstruction (L2)}& & \multicolumn{2}{c}{Samples (FID)} & & \multicolumn{2}{c}{Reconstruction (L2)}& & \multicolumn{2}{c}{Samples (FID)}\\
 \cline{2-3}\cline{5-6}\cline{8-9}\cline{11-12}
 & Train Err & Test Err & & $P_{Z}$ & $\hat{Q}_{Z;\phi}$ & & Train Err & Test Err & & $P_{Z}$ & $\hat{Q}_{Z;\phi}$\\
 \cline{2-6}\cline{8-12}
 VAE & 6.16 & 8.72 & & 23.20 & 23.03 & & 85.45 & 100.16 & & 58.84 & 54.19 \\
 VAE + IAF & 6.65 & 7.91 & & 29.30 &  & & 96.18 & 120.82 & & 50.73 & \\
 cWAE & \textbf{1.36} & 5.36 & & 30.08 & 8.22 & & 63.22 & 96.04 & & 61.29 & 47.09 \\
 cVCAE & \textbf{1.36} & \textbf{5.15} & &  & \textbf{7.68} & & \textbf{62.76} & \textbf{94.40} & &  & \textbf{43.14}\\
\end{tabular}
\end{table*}%

\begin{figure*}
	\centering
	\begin{subfigure}[t]{0.19\textwidth} \includegraphics[width=\textwidth]{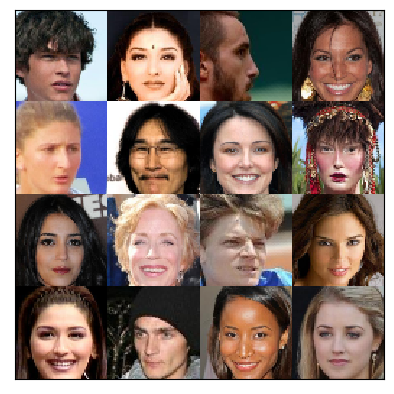} \caption{Original Test Set} \end{subfigure}
	\begin{subfigure}[t]{0.19\textwidth} \includegraphics[width=\textwidth]{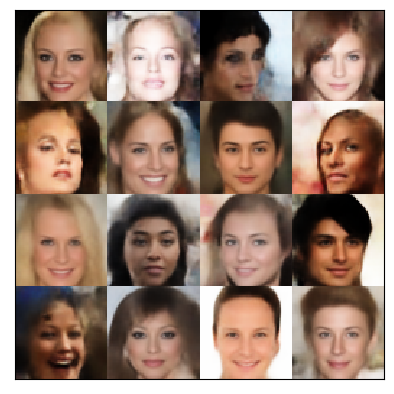} \caption{VCAE} \end{subfigure}
	\begin{subfigure}[t]{0.19\textwidth} \includegraphics[width=\textwidth]{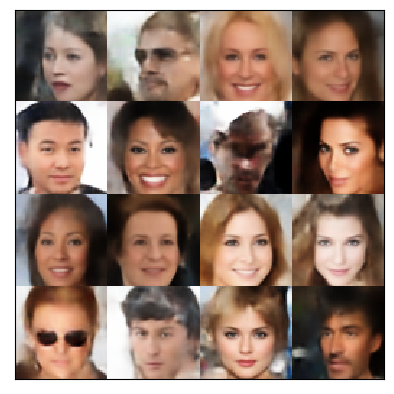} \caption{cWAE ($\hat{Q}_{Z;\phi}$)} \end{subfigure}
	\begin{subfigure}[t]{0.19\textwidth} \includegraphics[width=\textwidth]{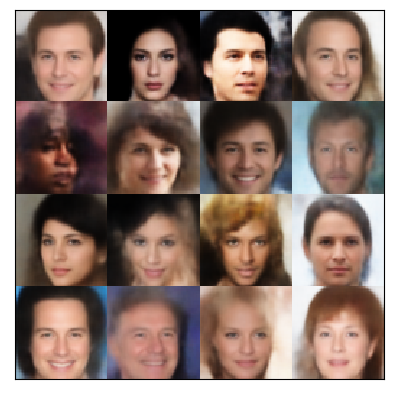} \caption{VAE + IAF} \end{subfigure}
	\begin{subfigure}[t]{0.19\textwidth} \includegraphics[width=\textwidth]{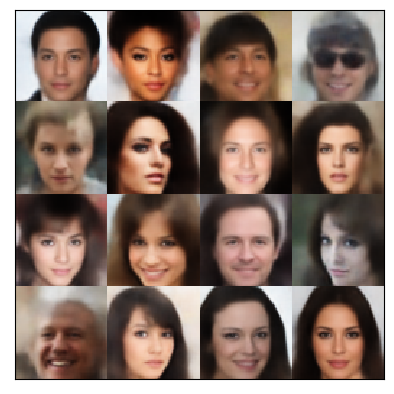} \caption{VAE ($\hat{Q}_{Z;\phi}$)} \end{subfigure} 
	
	\caption{Samples taken from generative models trained on the CelebA dataset.}
	\label{fig:celeba_image_results}
\end{figure*}

In this section, we compare the generative and reconstructive performance of VCAE, cWAE, VAE, and VAE + IAF on the two datasets MNIST and CelebA. These models are compared using training set errors, testing set errors, and Fréchet Inception Distance (FID) scores \citep{heusel2017gans}. When reporting FID scores for WAE and VAE, we investigate two situations, one where the assumed prior is used ($\ND(0, 1)$), and the other where the true latent distribution is learned using normalising flows. The latter allows for a fair comparison with VCAE. Sections \ref{sec:setup:mnist} \& \ref{sec:setup:celeba} of the appendix give the experimental setup for the MNIST and CelebA experiments, respectively. These sections contain results for different settings of $\lambda$ for both cWAE and VCAE.

Table \ref{tab:mnist_celeba_results} shows a quantitative comparison between VCAE, cWAE, VAE and VAE + IAF. The results show that both in the case of MNIST and CelebA, VCAE achieves the lowest testing set (mean square) error and the lowest FID score (lower is better) out of all other models. Figure \ref{fig:celeba_image_results} displays samples from each of the generative models trained on CelebA. A qualitative comparison of figure \ref{fig:celeba_image_results} shows that VCAE and WAE consistently produces higher quality samples than both VAE and VAE + IAF. In section \ref{sec:nn_analysis_on_celeba}, nearest neighbours (in the training set) of generations for these models trained on CelebA are displayed, showing that overfitting has not occurred. Figure \ref{fig:mnist_image_results}, which is found in section \ref{sec:aux_exp:generative_modelling_mnist} of the supplementary material, gives a qualitative comparison of the models trained on MNIST. Analysis of figure \ref{fig:mnist_image_results} yields the same results as were found for CelebA. 

\subsection{Disentanglement on Shapes3D}
\label{sec:vcae_disentanglemetn}
\begin{figure*}
	\centering
	\begin{subfigure}[t]{0.21\textwidth} \includegraphics[width=\textwidth]{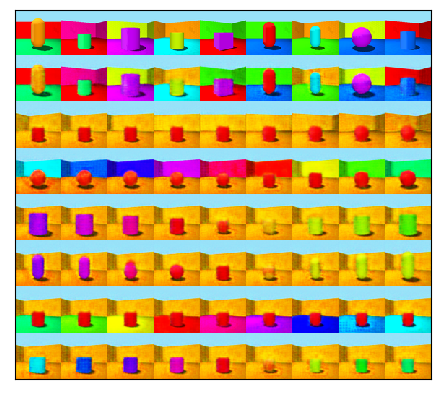} \caption{TC-VCAE} \end{subfigure}
	\begin{subfigure}[t]{0.21\textwidth} \includegraphics[width=\textwidth]{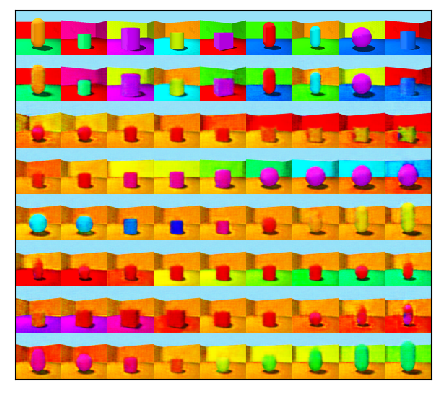} \caption{TC-cWAE} \end{subfigure}
	\begin{subfigure}[t]{0.21\textwidth} \includegraphics[width=\textwidth]{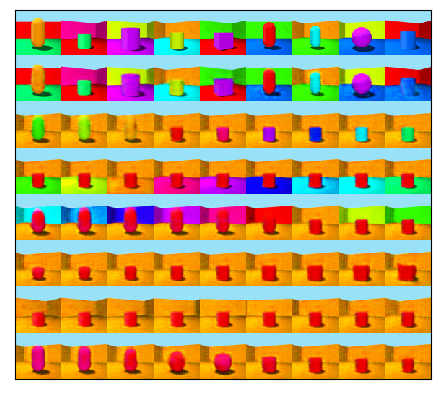} \caption{FactorVAE} \end{subfigure}
	
	\caption{Reconstructions and latent traversals models trained on 3D-Shapes. The first two rows show original images and their reconstructions, respectively. The remaining rows show a latent traversal of each feature.}
	\label{fig:3dshapes_latent_traversals}
\end{figure*}

In this section, we evaluate VCAE's ability to learn disentangled representations when equipped with a total correlation penalty term. This extended model is called TC-VCAE. As a reference system, we used cWAE equipped with the same total correlation penalty term, denoted TC-cWAE. Additionally, we compare against the state-of-the-art FactorVAE \citep{kim2018disentangling}. A complete description of the setup for these experiments is given in section \ref{sec:setup:disentanglement} of the supplementary material. We evaluate these models on the 3D-Shapes \citep{3dshapes18} dataset. The disentanglement metric used to give the following results is from \citep{kim2018disentangling}.

\begin{table}[h!]
	\centering
	\caption{Error and disentanglement score for best performing (score) TC-VCAE, TC-cWAE and FactorVAE}
	\label{tab:disentanglement_peformance}
	\begin{tabular}[t]{c ccc}
	 & TC-VCAE & TC-cWAE & FactorVAE\\
	\cline{2-4}
	Error & 3538 & 3531 & 3522 \\
	Score & 0.93 & 0.58 & 0.93 \\
	\end{tabular}
\end{table}

Table \ref{tab:disentanglement_peformance} shows the final errors and disentanglement scores for the best performing models. TC-VCAE and FactorVAE had very similar results, both outperforming TC-cWAE. Careful examination of the latent traversals in figure \ref{fig:3dshapes_latent_traversals} shows that both TC-VCAE and FactorVAE have captured room orientation, wall hue and floor hue. TC-cWAE fails to capture any. It is important to note that FactorVAE \citep{kim2018disentangling} achieves a higher disentanglement score, of 1.0 in \citep{kim2018disentangling} than what was obtained here. In \citep{locatello2018challenging}, the authors discuss that the initialisation is more important for successfully performing disentanglement than the hyper-parameter configuration. In section \ref{sec:aux_exp:disentanglement_var_analysis} of the appendix, we display results from five runs of these models, demonstrating the variation in performance.

In section \ref{sec:aux_exp:tcvcaep_latent_space_analysis} of the supplementary material, further analysis is given, in the form of histograms of the learned representations as well as a larger traversal. Examining the feature histograms shown in the supplementary material demonstrate that the features which are closer to Gaussian, do not correspond to a disentangled feature. In general, the underlying factors of generation for a dataset will not be Gaussian. Consequently, attempting to enforce a Gaussian, or any other shape on the aggregate posterior does not align with the task of disentanglement. 

The objectives of both TC-cWAE and FactorVAE are enforcing Gaussianity on the aggregated posterior, but are relying on Gaussianity not being enforced to perform disentanglement. On the other hand, VCAE does not enforce any shape on $Q_{Z;\phi}$. Consequently, where the objectives of TC-cWAE and FactorVAE are contradictory for disentanglement, VCAE's objectives are not.

\section{Conclusion}
Enforcing a desired prior distribution on the aggregate posterior of a generative model such as the Wasserstein Autoencoder (WAE) facilitates sampling. However, when stochastic encoders are used, this latent distribution constraint negatively affects the model's reconstruction and generative quality. This issue arises because achieving the minimum expected reconstruction error corresponds to a particular specification of the aggregate posterior $Q_{Z;\phi}$. By attempting to enforce $Q_{Z; \phi} = P_Z$, the optimisation process must find a compromise between $P_Z$ and the optimal specification. 

This paper proposed the Variance Constrained Autoencoder (VCAE), which only constrains the variance of the aggregate posterior rather than constraining its shape. This change in constraints means that the shape of the latent distribution is no longer regularised to conflict with the expected distortion. After training, the distribution of the aggregate posterior $Q_{Z;\phi}$ is not known. Therefore, to facilitate sampling from the trained VCAE, a chain of normalising flows can be optimised as a secondary stage, learning an invertible transform from a user-defined distribution $P_W$ to the aggregate posterior $Q_{Z;\phi}$. 

Our experimental results showed that VCAE outperforms VAE, VAE + IAF and cWAE in terms of reconstruction and generative performance on MNIST and CelebA. Moreover, VCAE is a more principled approach for learning disentangled representations as it does not assume a prior. Observing histograms of learned features from FactorVAE and TC-WAE demonstrated that a constraint on the latent distribution shape was counterproductive for disentanglement. Hence, providing evidence that the objectives of TC-cWAE and FactorVAE are contrary to disentanglement, whereas VCAE objective facilitates the task. When VCAE is equipped with a total correlation penalty term, it performs as well as FactorVAE for the task of disentanglement on 3D Shapes.

\subsubsection*{Acknowledgments}
This research was funded by GN.


\bibliography{ref} 
\bibliographystyle{icml2019}


\appendix
\section{Proofs}
\subsection{Proof of Proposition \ref{lem:jacobian_objective}}
\label{sec:prop_jacobian_objective_proof}
\begin{proof}
Let $\mu_\phi: X \rightarrow Z$ and $\mu_\theta: Z \rightarrow X$, the encoding and decoding operations respectively, be differentiable functions. Let $\epsilon \sim P_\epsilon$ be a zero-mean distribution with variance $\sigma_\epsilon^2$. For sufficiently small $\epsilon$, we can approximate the decoding operation as:
\begin{align*}
	\mu_\theta(\mu_\phi(x)) + J_{\mu_\theta}(\mu_\phi(x))\epsilon,
\end{align*}
where $J_{\mu_\theta}(\mu_\phi(x))$ is the Jacobian of the decoder at point $\mu_\phi(x)$. The reconstruction cost can therefore be written as:
\begin{align*}
	c(x, \hat{x}) &= \EX_\epsilon [ \| x - \hat{x} \|^2_2 ] \\
	&= \EX_\epsilon [ \| (x - \mu_\theta(\mu_\phi(x))) + J_{\mu_\theta}(\mu_\phi(x))\epsilon \|^2_2 ].
\end{align*}
To facilitate analysis of this objective, let $a = x - \mu_\theta(\mu_\phi(x))$ and $b = J_{\mu_\theta}(\mu_\phi(x))\epsilon$. Then:
\begin{align*}
	\EX_\epsilon[(a_i + b_i)^2] &= a_i^2 + \EX_\epsilon[b_i^2] + \EX_\epsilon[a_ib_i]
\end{align*}
$\sum_i a_i^2$ is simply the square error between $x$ and $\mu_\theta(\mu_\phi(x))$. Now consider the second term, $b_i$ is the dot product between the $i$th row of $J_{\mu_\theta}(\mu_\phi(x))$ (denote as $J_i$ for simplicity) and $\epsilon$:
\begin{align*}
	\EX_\epsilon[b_i^2] &= \EX_\epsilon[(J_i \cdot \epsilon)^2] \\
	&= \EX_\epsilon[(J_i \cdot \epsilon)(\epsilon^T \cdot J_i^T)] \\
	&= \sigma_\epsilon^2 \sum_k [J_k \cdot J_k^T]\\
	&= \sigma_\epsilon^2 \| J_{\mu_\theta}(\mu_\phi(x)) \|^2_2.
\end{align*}
The final term is zero because $P_\epsilon$ is a zero-mean distribution. 
Therefore, as required, we have:
\begin{multline*}
		\E_{\epsilon} [ \| x - \hat{x} \|^2_2] = \|x - \mu_\theta(\mu_\phi(x)) \|_2^2 \\+ \sigma_\epsilon^2 \| J_{\mu_\theta}(\mu_\phi(x)) \|^2_F.
	\end{multline*}
\end{proof}

\section{Experimental Setup}
\label{sec:exp_setup}

\begin{table*}
	\centering
	\caption{Description of neural network architectures used by the experiments in this paper. Conv refers to a 2D convolutional layer with parameters nf (number of filters), ks (kernel size) and s (stride). Similarly TConv refers to a 2D transpose convolutional layer with the same parameters. BN refers to a batch normalisation layer and ReLU refers to the non-linear Rectified Linear Unit activation function.}
	\label{tab:neural_network_archetchures}
	\begin{tabular}[t]{cc}
		\multicolumn{2}{c}{MNIST}\\
		\hline\\
		Encoder & Decoder \\
		\parbox{0.45\textwidth}{
		\begin{align*}
			&x \in \mathcal{R}^{28 \times 28}\\
			&\textrm{Conv[nf=128, ks=4, s=2], BN, ReLU}\\
			&\textrm{Conv[nf=256, ks=4, s=2], BN, ReLU}\\
			&\textrm{Conv[nf=512, ks=4, s=2], BN, ReLU}\\
			&\textrm{Conv[nf=1024, ks=4, s=2], BN, ReLU}\\
			&\textrm{Flatten}\\
			&\textrm{Dense[16]} \rightarrow \ \mu_\phi(x)
		\end{align*}}%
		& 
		\parbox{0.45\textwidth}{
			\begin{align*}
				&z \in \mathcal{R}^{16}\\
				&\textrm{Dense[}7 \times 7 \times 1024\textrm{]}\\
				&\textrm{Reshape[(-1, 7, 7, 1024)]}\\
				&\textrm{TConv[nf=512, ks=4, s=2]}, BN, ReLU\\
				&\textrm{TConv[nf=256, ks=4, s=2]}, BN, ReLU\\
				&\textrm{TConv[nf=1, ks=4, s=1]} \rightarrow \hat{x}
			\end{align*}
		}
		\\
		\multicolumn{2}{c}{CelebA}\\
		\hline\\
		Encoder & Decoder \\
		\parbox{0.45\textwidth}{
		\begin{align*}
			&x \in \mathcal{R}^{64 \times 64 \times 3}\\
			&\textrm{Conv[nf=128, ks=5, s=2], BN, ReLU}\\
			&\textrm{Conv[nf=256, ks=5, s=2], BN, ReLU}\\
			&\textrm{Conv[nf=512, ks=5, s=2], BN, ReLU}\\
			&\textrm{Conv[nf=1024, ks=5, s=2], BN, ReLU}\\
			&\textrm{Flatten}\\
			&\textrm{Dense[64]} \rightarrow \ \mu_\phi(x)
		\end{align*}}%
		& 
		\parbox{0.45\textwidth}{
			\begin{align*}
				&z \in \mathcal{R}^{64}\\
				&\textrm{Dense[}8 \times 8 \times 1024\textrm{]}\\
				&\textrm{Reshape[(-1, 8, 8, 1024)]}\\
				&\textrm{TConv[nf=512, ks=5, s=2]}, BN, ReLU\\
				&\textrm{TConv[nf=256, ks=5, s=2]}, BN, ReLU\\
				&\textrm{TConv[nf=128, ks=5, s=2]}, BN, ReLU\\
				&\textrm{TConv[nf=3, ks=5, s=1]} \rightarrow \hat{x}
			\end{align*}
		}
		\\
		\multicolumn{2}{c}{3D Shapes}\\
		\hline\\
		Encoder & Decoder \\
		\parbox{0.45\textwidth}{
		\begin{align*}
			&x \in \mathcal{R}^{64 \times 64 \times 3}\\
			&\textrm{Conv[nf=32, ks=4, s=2], ReLU}\\
			&\textrm{Conv[nf=32, ks=4, s=2], ReLU}\\
			&\textrm{Conv[nf=64, ks=4, s=2], ReLU}\\
			&\textrm{Conv[nf=64, ks=4, s=2], ReLU}\\
			&\textrm{Flatten}\\
			&\textrm{Dense[256]}\\
			&\textrm{Dense[6]} \rightarrow \ \mu_\phi(x)
		\end{align*}}%
		& 
		\parbox{0.45\textwidth}{
			\begin{align*}
				&z \in \mathcal{R}^{6}\\
				&\textrm{Dense[256], ReLU}\\
				&\textrm{Dense[} 4 \times 4 \times 64 \textrm{], ReLU}\\
				&\textrm{Reshape[(-1, 4, 4, 64)]}\\
				&\textrm{TConv[nf=64, ks=4, s=2]}, ReLU\\
				&\textrm{TConv[nf=32, ks=4, s=2]}, ReLU\\
				&\textrm{TConv[nf=32, ks=4, s=2]}, ReLU\\
				&\textrm{TConv[nf=3, ks=4, s=2]} \rightarrow \hat{x}
			\end{align*}
		}
		\\
	\end{tabular}
\end{table*}

\begin{table*}
	\centering
	\caption{Description of normalising flow architectures used by the experiments in this paper. MAF stands for Masked Autoregressive Flow \cite{papamakarios2017masked}. There is a Permutation layer in-between each MAF as this improves performance \cite{papamakarios2017masked}.}
	\label{tab:normalising_flow_archetchures}
	\begin{tabular}[t]{cc}
		MNIST & CelebA \\
		\hline\\
		\parbox{0.45\textwidth}{
		\begin{align*}
			&z \in \mathcal{R}^{16}\\
			&\textrm{MAF[64, 64]}, \ 5 \times \{ \textrm{Permutation}, \textrm{MAF[64, 64]} \}\\
			&\textrm{LinearScale}, \textrm{LinearShift} \rightarrow w \sim \ND(0, 1) \\
		\end{align*}}%
		& 
		\parbox{0.45\textwidth}{
			\begin{align*}
				&z \in \mathcal{R}^{64}\\
				&\textrm{RealNVP[256, 256, 256]}, \ 7 \times \{\textrm{Permutation}, \\ &\textrm{RealNVP[256, 256, 256]} \}\\
				& \rightarrow w \sim \ND(0, 1) \\
			\end{align*}
		}
		\\
	\end{tabular}
\end{table*}

In this section, we give a complete description of the setups for all experiments that were run in section \ref{sec:experiments}. The models evaluated are VAE, VAE + IAF, WAE and VCAE. We chose VAE + IAF over VAE + NF because it uses a more powerful class of functions which perform better in higher dimensions.

The implementation of VCAE follows Algorithm \ref{alg:vcae_algorythm}, and the implementation of WAE, VAE and VAE + IAF follow from their respective papers. In the case of WAE, we only consider the MMD-WAE, where the Maximum Mean Discrepancy (MMD) \cite{gretton2012kernel} is implemented using the inverse multi-quadratic (IMQ) kernel, $k(x,y) = C/(C + \| x - y \|^2_2)$, as used in \cite{tolstikhin2017wasserstein}, where kernel parameter $C$ is given by $2 \cdot z_{dim} \cdot \sigma^2_z$. For each given experiment, all models use the same neural network structure, the only exception being in the case of VAE and VAE + IAF, where the encoder outputs additional information. In the case of VAE, a final linear layer of the encoder outputs two vectors of length $z_{dim}$, $\mu_\phi(x)$ and $\Sigma_\phi(x)$, representing the mean and standard deviation of $Q_{Z|X;\phi}$. Additionally, in the case of VAE + IAF, the encoder outputs another vector of $z_{dim}$ length (for a total of three) which is provided as an additional input to the normalising flows, as per the standard implementation \cite{kingma2016improved}.

We note that table \ref{tab:neural_network_archetchures} describes the neural network architectures used in each experiment. The encoder/decoder structure for MNIST and CelebA follows that of \cite{tolstikhin2017wasserstein}, and the structure for 3D shapes follows that of \cite{kim2018disentangling}.

\begin{algorithm}
\centering
	\begin{algorithmic}
	\STATE {\bfseries Input signal} data $x_i$
   \STATE {\bfseries Output} Optimised parameters $\theta^*, \phi^*$
   \STATE set noise distribution $P_{\epsilon; \phi}$
   \STATE set weight $\lambda$ 
   \STATE set paramater $v$
   \STATE initialise parameters $\theta, \phi$
   \FOR{each minibatch $l\in\mathcal{L}_m$}
   \STATE $\bar{x}_l \leftarrow $ current minibatch
   \STATE $\bar{z} \leftarrow \mu_\phi(\bar{x}_l) + \bar{\epsilon}, \,\,\bar{\epsilon} \sim P_{\epsilon; \phi}$  \hspace{1em} \% encoder
   \STATE $\hat{x} \leftarrow \mu_\theta(\bar{z})$  \hspace{1em} \% decoder
   \STATE $L \leftarrow MSE(\hat{x}, \bar{x}_l) + \lambda  \, |\text{var}(\bar{z}) - v|$
   \STATE $(\theta, \phi) \leftarrow$ +Adam update of $\theta, \phi$ to minimise $L$
   \ENDFOR
   \STATE
   \STATE initialise $f_1$, ..., $f_m$
   \STATE $q_\psi(w) \leftarrow p(w) \cdot \prod_{i=1}^m | \det \frac{\partial f_i(w_{i-1})}{\partial w_{i-1}}|^{-1}$
   \FOR{each minibatch $l\in\mathcal{L}_n$}
   \STATE $\bar{x}_l \leftarrow $ current minibatch
   \STATE $\bar{z} \leftarrow \mu_\phi(\bar{x}_l) + \bar{\epsilon}, \,\,\bar{\epsilon} \sim P_{\epsilon; \phi}$  \hspace{1em} \% encoder
   \STATE $L_{nf} \leftarrow -\frac{1}{|\bar{x}_l|} \sum_{j=1}^{|\bar{x}_l|} [\log q_\psi(\bar{z}_j)]$
   \STATE $\psi \leftarrow$ +Adam update of $\psi$ to minimise $L_{nf}$
   \ENDFOR
	\end{algorithmic}
	\caption{VCAE algorithm with the assumption that $P_{X|Z;\theta}$ is a symmetric Gaussian. Let $\mathcal{L}_m$ and $\mathcal{L}_n$ be the number of minibatches used to train the encoder/decoder and normalising flows respectively.}
   \label{alg:vcae_algorythm}
\end{algorithm}

\subsection{Overfitting}
\label{sec:setup:overfitting}
For these experiments, we used the encoder/decoder structure outlined in table \ref{tab:neural_network_archetchures} under the MNIST heading. To outline the effect of overfitting we train on a reduced version of MNIST dataset (we denote this dataset ReducedMNIST), which consists of a 600 sample subset of the training data. For VCAE we select $\lambda_{VCAE} = 2$, and $v = 2$. In the case of WAE, we select $\lambda_{WAE} = 100$, and $P_Z = \ND(0, I_{2})$. A number of experiments were run with different choices for $P_\epsilon$ (for VCAE and WAE), these are $\ND(0, 0.2 \cdot I_{2})$, $\ND(0, 0.1 \cdot I_{2})$ and $\epsilon = 0$ (no noise).

During the training of these models, the Adam optimiser was used with an initial learning rate of $1\times 10^{-4}$ and no learning rate schedule. A batch size of 200 was used in all cases.

\subsection{Latent Space Analysis}
\label{sec:setup:latent_space_analysis}
For these experiments, we used the encoder/decoder structure outlined in table \ref{tab:neural_network_archetchures} under the MNIST heading. For VCAE we select $\lambda_{VCAE} = 1$, and $v = 2$. In the case of WAE, we select $\lambda_{WAE} = 100$, and $P_Z = \ND(0, I_{2})$. We selected $P_\epsilon = \ND(0, 0.01 \cdot I_{2})$.

During the training of these models, the Adam optimiser was used with an initial learning rate of $1\times 10^{-4}$ and no learning rate schedule. A batch size of 200 was used in all cases.

\subsection{MNIST}
\label{sec:setup:mnist}
In this section, we describe the setup for the experiments comparing the generative and reconstructive quality of VCAE, cWAE, VAE and VAE+IAF on MNIST. 

\subsubsection{Model Setup}
For this experiment we use the encoder/decoder setup described under the MNSIT heading in table \ref{tab:neural_network_archetchures}. For all experiments we chose $z_{dim} = 16$. In the case of VCAE and cWAE, chose $P_\epsilon = \ND(0, 0.05 \cdot I_{16})$. The normalising flow architecture used for these experiments is given in table \ref{tab:normalising_flow_archetchures} under the MNIST heading.

During training, we used a batch size of 100 in all cases. For VCAE, WAE and VAE we used an Adam optimiser with an initial learning rate of $1\times 10^{-3}$, for VAE + IAF we used an Adam optimiser with the initial learning rate $1\times 10^{-4}$. For all experiments, the learning rate schedule was the same: after 30 epochs, cut the learning rate in half; after 50 epochs, reduce the learning rate by a factor of five. For VAE, WAE and VCAE, the encoder and decoder are trained for 100 epochs. VAE + IAF is trained for 200 epochs as it has a lower learning rate (because it does not converge for a larger learning rate) and extra parameters (the normalising flows).

When training the normalising flows as a secondary stage, we train for 100 epochs of the dataset using the Adam optimiser with an initial learning rate of $1\times 10^{-3}$ and no learning rate schedule. The batch size used was 100.

\subsubsection{Hyper-paramater Selection}
For VCAE and WAE the hyper-parameter $\lambda$ must be chosen, this parameter controls the tradeoff between minimising output distortion and enforcing the variance or distribution constraint. To ensure a fair comparison, we would like to find a parameter setting that sufficiently enforces the desired constraint while performing as optimally as possible. In tables \ref{tab:vcae_lambda_explore_mnist} \& \ref{tab:cwae_lambda_explore_mnist} we explore various settings of $\lambda$ for VCAE and WAE respectively. Under both constraints, we should find that the sum of variances in the latent dimension should equal $16$, hence, we choose the setting of $\lambda$ for which the summed latent variance is approximately $16$. The hyper-parameters for VCAE are: 1) $\lambda_{VCAE} = 0.1$; 2) $v = z_{dim} = 16$. The hyper-parameters for WAE are: 1) $\lambda_{WAE} = 50$; 2) $P_Z = \ND(0, I_{16})$.

\begin{table}[H]
	\centering
	\begin{tabular}[t]{c cccc}
		\toprule
		$\lambda$ & Train & Test & & $\Sigma\ var(z_i)$ \\
		\cline{2-5}\\
		0.05 & 1.35 & 5.19 & & 17.29 \\
		\textbf{0.1} & \textbf{1.36} & \textbf{5.15} & & \textbf{16.05} \\
		0.5 & 1.47 & 5.01 & & 16.40 \\
		1.0 & 1.52 & 5.60 & & 16.86 \\
		1.5 & 1.56 & 5.76 & & 16.70 \\
		\bottomrule
	\end{tabular}
	\caption{Exploration of VCAE performance on MNIST ($z_{dim} = 16$) with different settings of the hyper-parameter $\lambda$.}
	\label{tab:vcae_lambda_explore_mnist}
\end{table}

\begin{table}[H]
	\centering
	\begin{tabular}[t]{c ccccc}
		\toprule
		& \multicolumn{2}{c}{Error} & & \multicolumn{1}{c}{FID} & \\
		\cline{2-3} \cline{5-5}
		$\lambda$ & Train & Test & & $P_Z$ & $\Sigma\ var(z_i)$ \\
		\cline{2-6}\\
		5 & 1.24 & 5.17 & & 27.27 & 23.45 \\
		10 & 1.27 & 5.15 & & 26.26 & 19.85 \\
		25 & 1.28 & 5.33 & & 27.38 & 17.55 \\
		\textbf{50} & \textbf{1.36} & \textbf{5.56} & & \textbf{30.08} & \textbf{16.51} \\
		75 & 1.43 & 6.023 & & 33.46 & 15.78 \\
		100 & 1.48 & 6.26 & & 33.57 & 15.67 \\
		150 & 1.53 & 6.67 & & 34.27 & 15.55 \\
		\bottomrule
	\end{tabular}
	\caption{Exploration of cWAE performance on MNIST ($z_{dim} = 16$) with different settings of the hyper-parameter $\lambda$.}
	\label{tab:cwae_lambda_explore_mnist}
\end{table}

\subsection{CelebA}
\label{sec:setup:celeba}
In this section, we describe the experimental setup for experiments performed on CelebA. We used a pre-processsed version of the CelebA dataset, obtained via the following steps:
\begin{itemize}
	\item Take a $140\times 140$ pixel centre crop of each image.
	\item Down scale each cropped image to $64\times 64$ pixels.
\end{itemize}
The CelebA data set is pre-processed in the same way for the experiments conducted in \cite{tolstikhin2017wasserstein}.

\subsubsection{Model Setup}
For these experiment we used the encoder/decoder setup described in table \ref{tab:neural_network_archetchures} under CelebA. For all experiments we chose $z_{dim} = 64$. In the case of VCAE and cWAE, chose $P_\epsilon = \ND(0, 0.05 \cdot I_{64})$. The normalising flow architecture used for these experiments is given in table \ref{tab:normalising_flow_archetchures} under the CelebA heading.

During training, we used a batch size of 100 in all cases. For VCAE, WAE and VAE we used an Adam optimiser with an initial learning rate of $1\times 10^{-3}$, for VAE + IAF we used an Adam optimiser with the initial learning rate $1\times 10^{-4}$. For all experiments, the learning rate schedule was the same: after 30 epochs, cut the learning rate in half; after 50 epochs, reduce the learning rate by a factor of five. For VAE, WAE and VCAE, the encoder and decoder are trained for 70 epochs. VAE + IAF is trained for 140 epochs as it has a lower learning rate (because it does not converge for a larger learning rate) and extra parameters (the normalising flows).

When training the normalising flows as a secondary stage, we train for 100 epochs of the dataset using the Adam optimiser with an initial learning rate of $1\times 10^{-3}$ and no learning rate schedule. The batch size used was 100.

\subsubsection{Hyper-paramater Selection}
For these experiments we must also select $lambda_{VCAE}$ and $\lambda_{WAE}$ which control the tradeoff between minimising the reconstruction cost and enforcing the constraint on the aggregate posterior $Q_{Z;\phi}$. To ensure a fair comparison, we would like to find a parameter setting that sufficiently enforces the desired constraint while performing as optimally as possible. In tables \ref{tab:vcae_lambda_explore_celeba} \& \ref{tab:cwae_lambda_explore_celeba} we explore various settings of $\lambda$ for VCAE and WAE respectively. Under both constraints, we should find that the sum of variances in the latent dimension should equal $64$, hence, we choose the setting of $\lambda$ for which the summed latent variance is approximately $16$. The hyper-parameters for VCAE are: 1) $\lambda_{VCAE} = 0.5$; 2) $v = z_{dim} = 64$. The hyper-parameters for WAE are: 1) $\lambda_{WAE} = 750$; 2) $P_Z = \ND(0, I_{64})$.

\begin{table}[H]
	\centering
	\begin{tabular}[t]{c cccc}
		\toprule
		$\lambda$ & Train & Test & & $\Sigma\ var(z_i)$ \\
		\cline{2-5}\\
		0.05 & 59.98 & 97.82 & & 65.24 \\
		0.1 & 60.26 & 96.63 & & 64.64 \\
		\textbf{0.5} & \textbf{62.77} & \textbf{94.40} & & \textbf{63.47} \\
		\bottomrule
	\end{tabular}
	\caption{Exploration of VCAE performance on CelebA ($z_{dim} = 64$) with different settings of the hyper-parameter $\lambda$.}
	\label{tab:vcae_lambda_explore_celeba}
\end{table}

\begin{table}[H]
	\centering
	\begin{tabular}[t]{c ccccc}
		\toprule
		& \multicolumn{2}{c}{Error} & & \multicolumn{1}{c}{FID} & \\
		\cline{2-3} \cline{5-5}
		$\lambda$ & Train & Test & & $P_Z$ & $\Sigma\ var(z_i)$ \\
		\cline{2-6}\\
		100 & 60.63 & 97.78 & & 61.99 & 73.52 \\
		250 & 61.59 & 99.25 & & 61.70 & 68.32 \\
		500 & 61.24 & 98.81 & & 59.05 & 66.93 \\
		\textbf{750} & \textbf{63.22} & \textbf{96.04} & & \textbf{61.29} & \textbf{65.70} \\
		1000 & 65.94 & 98.22 & & 64.63  & 65.72 \\
		\bottomrule
	\end{tabular}
	\caption{Exploration of cWAE performance on CelebA ($z_{dim} = 64$) with different settings of the hyper-parameter $\lambda$.}
	\label{tab:cwae_lambda_explore_celeba}
\end{table}

\subsection{Disentanglement on 3D Shapes}
\label{sec:setup:disentanglement}
For these experiments, we follow the setup described in \cite{kim2018disentangling}. However, to make this paper self-contained, we will reiterate the setup here. The encoder/decoder setup is described in table \ref{tab:neural_network_archetchures} under the 3D shapes heading. The same structure is used for all experiments except for the implementation differences outlined at the start of this section. 

The implementation of the total correlation (TC) penalty term requires an additional discriminator network which consists of six fully-connected layers, each with 1000 hidden units and used the LeakyReLU ($\alpha = 0.2$) activation. The discriminator network outputs two logits.

Each model was trained for a total of $5 \times 10^5$ batches, with a batch size of 64. Six latent dimensions were used. Adding the total correlation penalty introduces another hyper-parameter $\gamma$, which controls how strongly the constraint is enforced. For all experiments, we choose $\gamma = 7$, as this was reported to be the optimal setting for FactorVAE \cite{kim2018disentangling}. For VCAE we choose $\lambda_{VCAE} = 1$ and for cWAE we choose $\lambda_{WAE} = 2500$. For both VCAE and cWAE we choose $P_\epsilon = \ND(0, 0.1 \cdot I_{6})$.

\section{Auxiliary Experiments}

\subsection{Toy MNIST Example}
\label{sec:aux:toy_mnist_example}
In this section we give further figures to complement the analysis given in section \ref{sec:toy_experiment}. Figure \ref{fig:mnist_image_results} gives histograms of the latent features learned by cWAE and VCAE for the MNIST toy experiment. In this experiment, we chose $P_\epsilon = \ND(0.03, I)$. We progressively increased the regularisation parameter for cWAE, resulting in $\lambda_{cWAE} = 3000$. Increasing $\lambda_{cWAE}$ resulted in a model that could not train.

\begin{figure*}
	\centering
	\begin{subfigure}[t]{0.22\textwidth} \includegraphics[width=\textwidth]{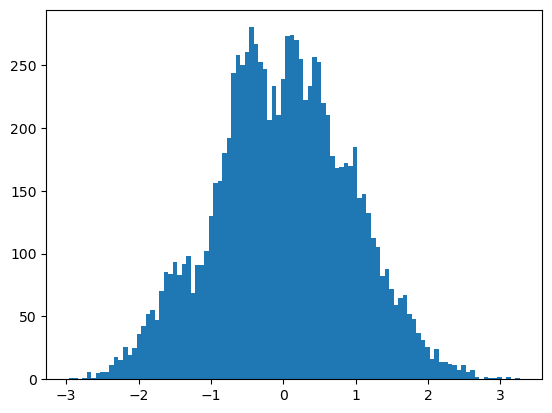} \caption{cWAE Feature 1} \end{subfigure}
	\begin{subfigure}[t]{0.22\textwidth} \includegraphics[width=\textwidth]{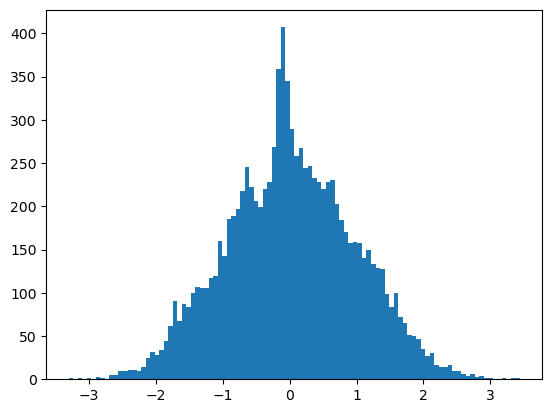} \caption{cWAE Feature 2} \end{subfigure}
	\hspace{0.08\textwidth}
	\begin{subfigure}[t]{0.22\textwidth} \includegraphics[width=\textwidth]{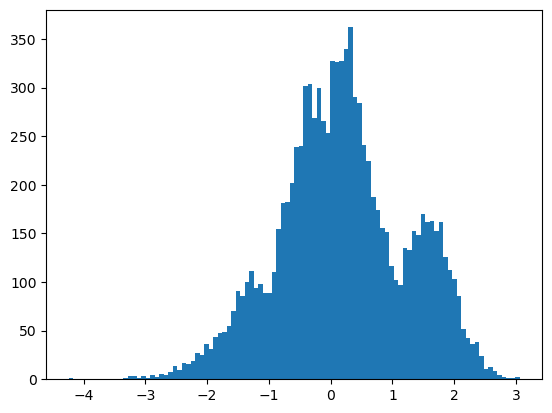} \caption{VCAE Feature 1} \end{subfigure}
	\begin{subfigure}[t]{0.22\textwidth} \includegraphics[width=\textwidth]{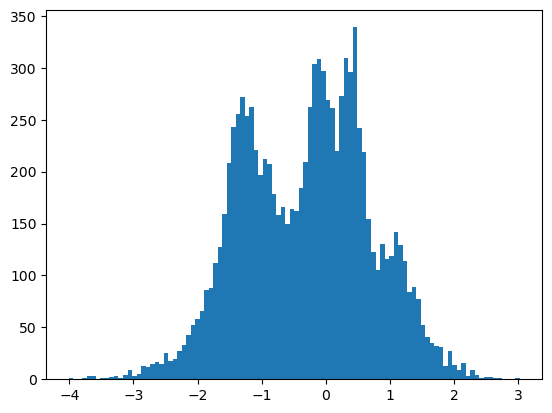} \caption{VCAE Feature 2} \end{subfigure}
	
	\caption{Comparison of feature histograms for cWAE and VCAE in the case of the toy MNIST experiment.}
	\label{fig:mnist_image_results}
\end{figure*}

\subsection{Generative Modeling on MNIST}
\label{sec:aux_exp:generative_modelling_mnist}
\begin{figure*}
	\centering
	\begin{subfigure}[t]{0.19\textwidth} \includegraphics[width=\textwidth]{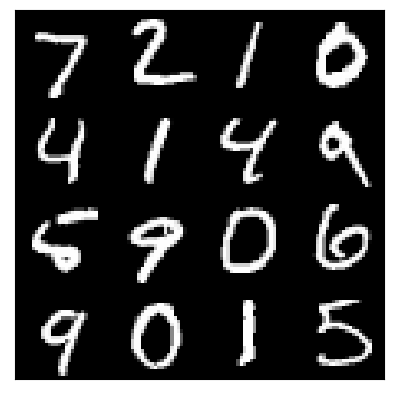} \caption{Original Test Set} \end{subfigure}
	\begin{subfigure}[t]{0.19\textwidth} \includegraphics[width=\textwidth]{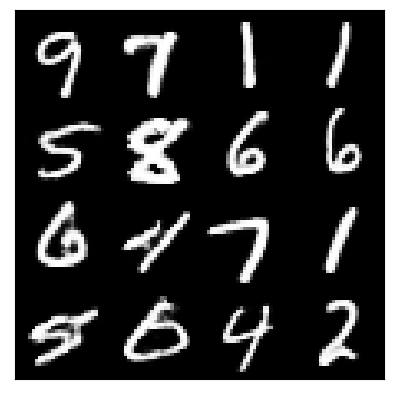} \caption{VCAE} \end{subfigure}
	\begin{subfigure}[t]{0.19\textwidth} \includegraphics[width=\textwidth]{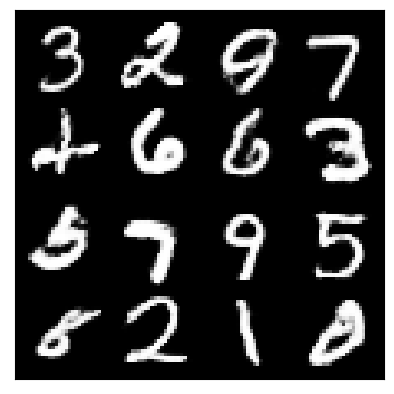} \caption{cWAE ($\hat{Q}_{Z;\phi}$)} \end{subfigure}
	\begin{subfigure}[t]{0.19\textwidth} \includegraphics[width=\textwidth]{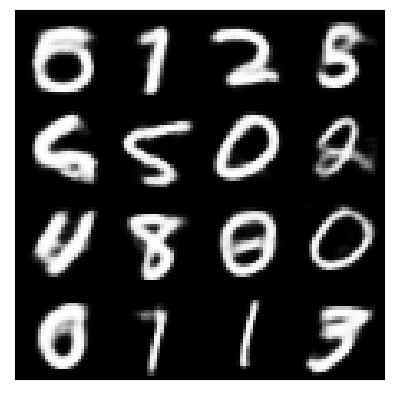} \caption{VAE + IAF} \end{subfigure}
	\begin{subfigure}[t]{0.19\textwidth} \includegraphics[width=\textwidth]{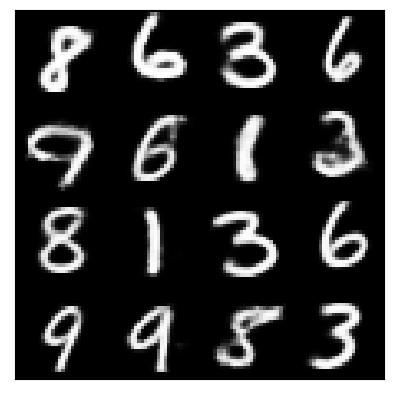} \caption{VAE ($\hat{Q}_{Z;\phi}$)} \end{subfigure} 
	
	\caption{Samples taken from generative models trained on the MNIST dataset.}
	\label{fig:mnist_image_results}
\end{figure*}

\subsection{Nearest Neighbour Analysis on CelebA}
\label{sec:nn_analysis_on_celeba}
\begin{figure*}
	\centering
	\begin{subfigure}[t]{0.19\textwidth} \includegraphics[width=\textwidth]{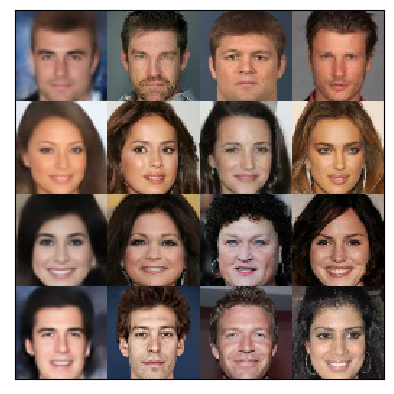} \caption{VCAE} \end{subfigure}
	\begin{subfigure}[t]{0.19\textwidth} \includegraphics[width=\textwidth]{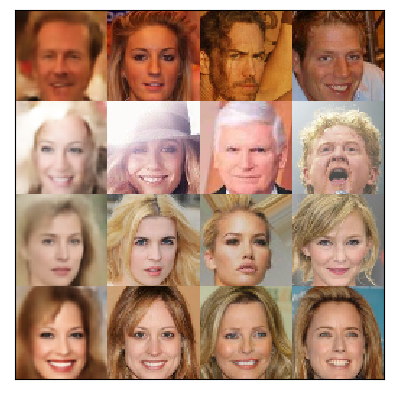} \caption{cWAE} \end{subfigure}
	\begin{subfigure}[t]{0.19\textwidth} \includegraphics[width=\textwidth]{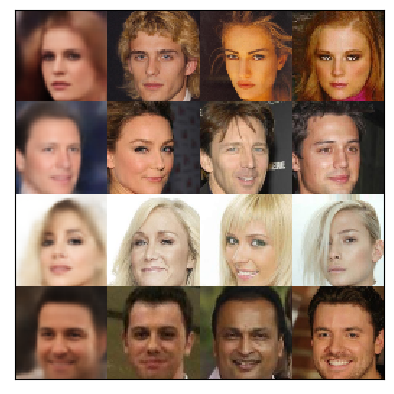} \caption{VAE} \end{subfigure}
	\begin{subfigure}[t]{0.19\textwidth} \includegraphics[width=\textwidth]{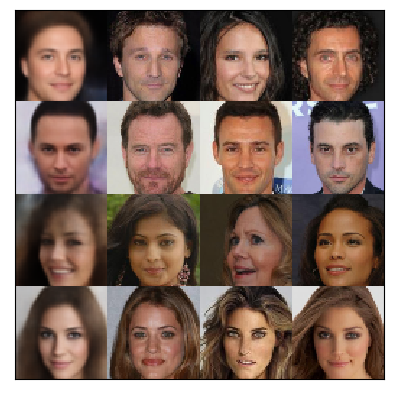} \caption{VAE + IAF} \end{subfigure}
	
	\caption{Training set nearest neighbours of generations from four models trained on CelebA. VCAE, cWAE and VAE have had their aggregate posteriors approximated with normalising flows.}
	\label{fig:mnist_image_results}
\end{figure*}

\subsection{Latent Space Analysis}
\label{sec:aux_exp:latent_space_analysis}
\begin{figure*}[htpb]
	\centering
	\begin{subfigure}{0.24\textwidth}
		\includegraphics[width=\textwidth]{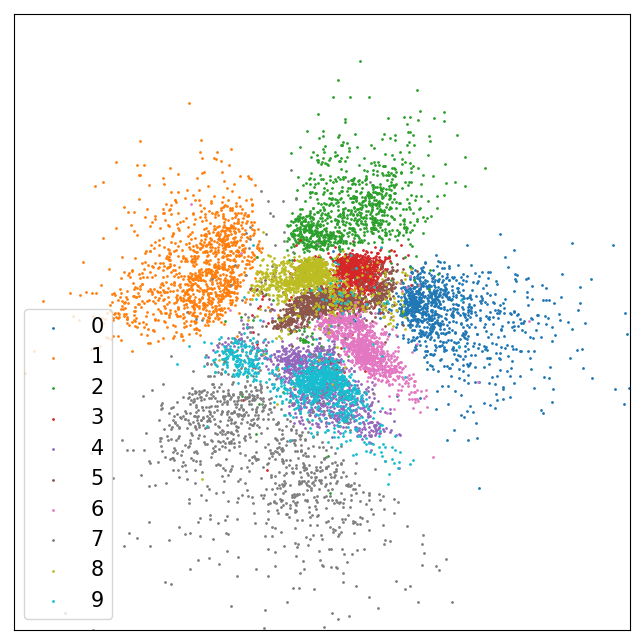}
		\caption{VCAE: $Q_{Z;\phi}$}
	\end{subfigure}
	\begin{subfigure}{0.24\textwidth}
		\includegraphics[width=\textwidth]{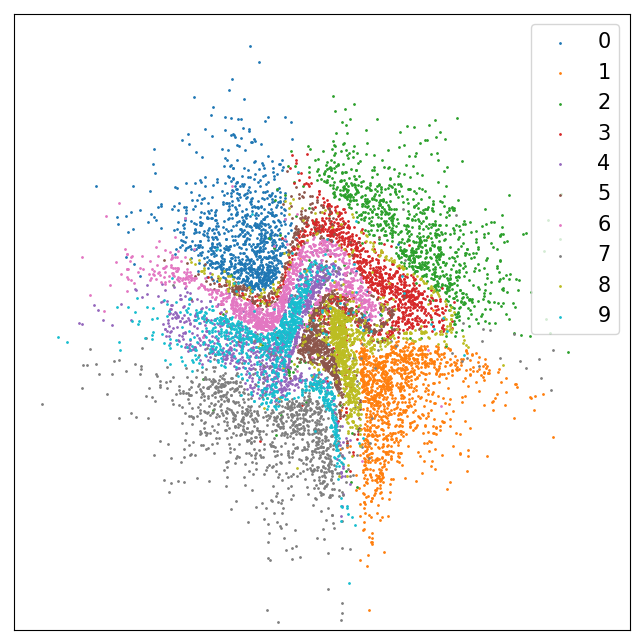}
		\caption{VCAE: $P_W$}
	\end{subfigure}
	\begin{subfigure}{0.24\textwidth}
		\includegraphics[width=\textwidth]{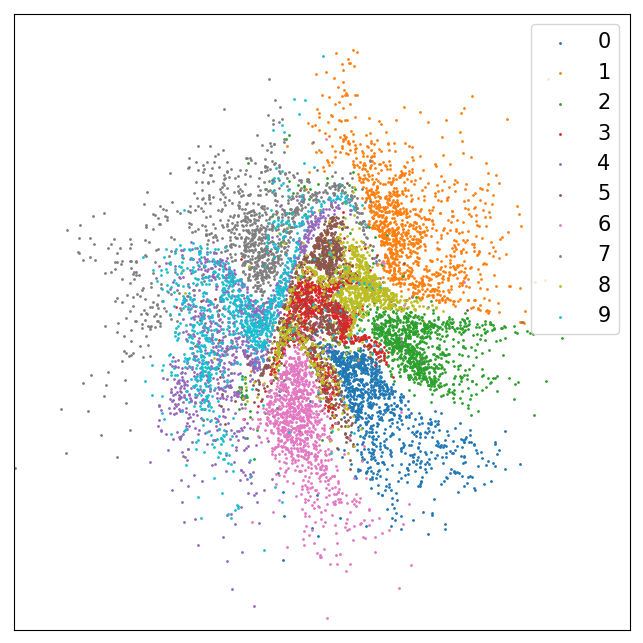}
		\caption{WAE: $P_W$}
	\end{subfigure}
	\begin{subfigure}{0.24\textwidth}
		\includegraphics[width=\textwidth]{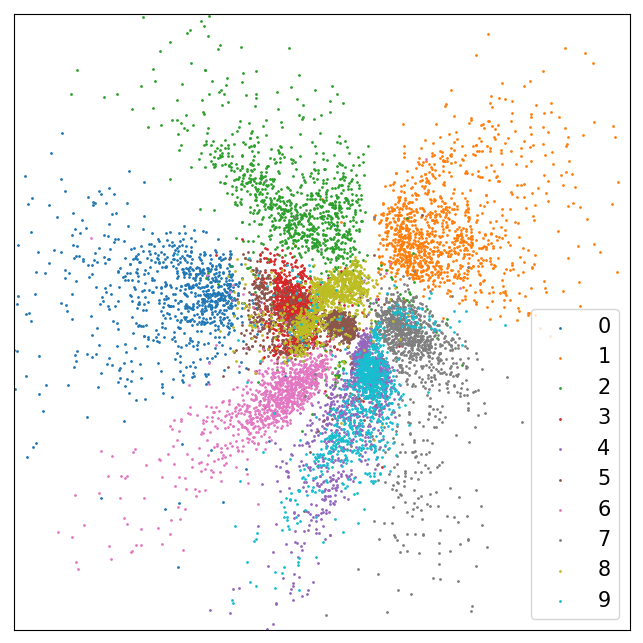}
		\caption{WAE: $Q_{Z;\phi}$}
	\end{subfigure}
	
	\caption{Figures showing $Q_{Z;\phi}$ and $NF^{-1}(Q_{Z;\phi})$ for a VCAE and WAE.}
	\label{fig:mnist_2d_latent_embeddings}
\end{figure*}

In this section, we develop a better understanding of the learned latent distributions and the effects that applying a normalising flow transform has on them. We trained VCAE and WAE on MNIST with a two-dimensional latent space, as this facilitates visualisation of $Q_{Z;\phi}$. Additionally, we train a chain of normalising flows to transform $P_W$ (selected to be a unit Gaussian) into $Q_{Z;\phi}$ for both models, we can then display the distribution $\hat{Q}_{Z;\phi}$ inversely mapped through the normalising flow denoted $NF^{-1}(Q_{Z;\phi}) = P_W$.

Figure \ref{fig:mnist_2d_latent_embeddings} shows four latent embedding plots, showing both $\hat{Q}{Z;\phi}$ and $P_W$ for VCAE and WAE. We observed that for both VCAE and cWAE $\hat{Q}_{Z;\phi}$ contains gaps between the different classes, in these regions, the decoder behaviour is undefined. The Gaussian representation for both models does not include these gaps, meaning we avoid sampling from these undefined regions. 

\subsection{Variability Analysis}
Table \ref{tab:disentanglement_run_results} shows the error and scores from five runs of FactorVAE, TC-VCAE and TC-WAE. The results demonstrate how varied performance was for different random initialisations.

\label{sec:aux_exp:disentanglement_var_analysis}
\begin{table*}[]
	\centering
	\caption{Error and disentanglement scores obtained from five runs of FactorVAE, TC-VCAE and TC-WAE.}
	\label{tab:disentanglement_run_results}
	\begin{tabular}[t]{c ccccc}
	\toprule
	 & Run 1 & Run 2 & Run 3 & Run 4 & Run 5\\
	\cline{2-6}\\
	& \multicolumn{5}{c}{FactorVAE}\\
	\cline{2-6}
	Error  & 3518.21 & 3517.20 & 3507.24 & 3521.96 & 3508.93\\
	D-Score & 0.78 & 0.90 & 0.60 & 0.93 & 0.67 \\
	
	\cline{2-6}\\
	& \multicolumn{5}{c}{TC-WAE}\\
	\cline{2-6}
	Error  & 3518.62 & 3516.12 & 3517.69 & 3513.09 & 3522.99 \\
	D-Score & 0.53 & 0.55 & 0.56 & 0.58 & 0.49 \\
	
	\cline{2-6}\\
	& \multicolumn{5}{c}{TC-VCAE}\\
	\cline{2-6}
	Error  & 3538.53 & 3578.88 & 3549.36 & 3529.74 & 3538.55\\
	D-Score & 0.93 & 0.69 & 0.64 & 0.91 & 0.87 \\
	\bottomrule
	\end{tabular}
\end{table*}

\subsection{Traversal and Latent Space Analysis}
\label{sec:aux_exp:tcvcaep_latent_space_analysis}
\begin{figure*}
	\centering
	\begin{subfigure}[t]{0.90\textwidth} 
	\includegraphics[width=\textwidth]{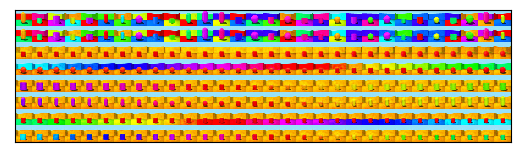} 
		\caption{TC-VCAE} 
		\label{fig:long_traversals_tcvcae}
	\end{subfigure}
	\begin{subfigure}[t]{0.90\textwidth} 
	\includegraphics[width=\textwidth]{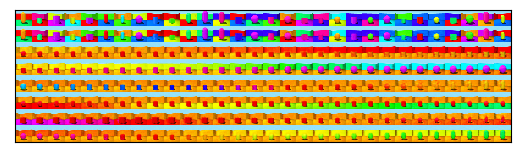} 
		\caption{TC-WAE} 
		\label{fig:long_traversals_tcwae}
	\end{subfigure}
	\begin{subfigure}[t]{0.90\textwidth} 
	\includegraphics[width=\textwidth]{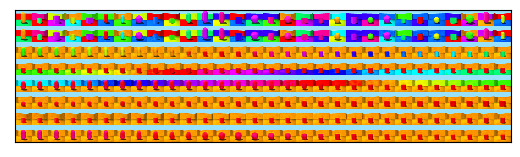} 
		\caption{FactorVAE} 
		\label{fig:long_traversals_factorvae}
	\end{subfigure}
	
	\caption{A traversal of the latent features from the best performing TC-VCAE, TC-WAE and FactorVAE for disentanglement on 3D Shapes.}
	\label{fig:long_traversals}
\end{figure*}

\begin{figure*}
	\centering
	\begin{subfigure}[t]{0.30\textwidth} \includegraphics[width=\textwidth]{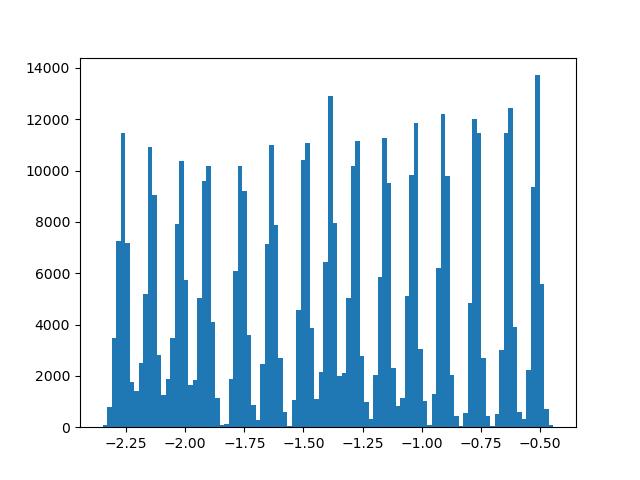} \caption{Feature 1 (Orientation)} \end{subfigure}
	\begin{subfigure}[t]{0.30\textwidth} \includegraphics[width=\textwidth]{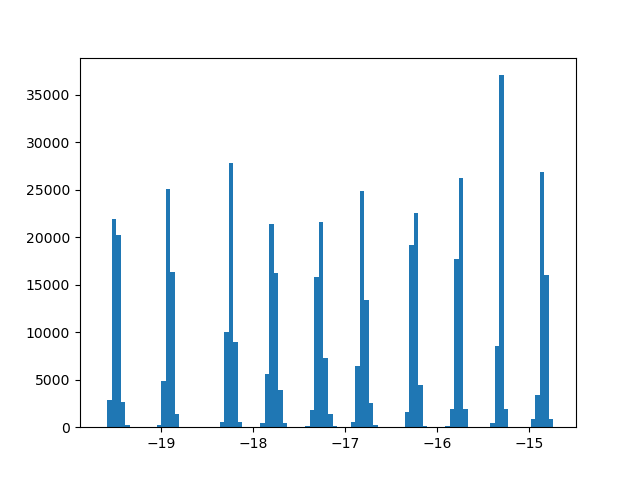} \caption{Feature 2 (Wall Hue)} \end{subfigure}
	\begin{subfigure}[t]{0.30\textwidth} \includegraphics[width=\textwidth]{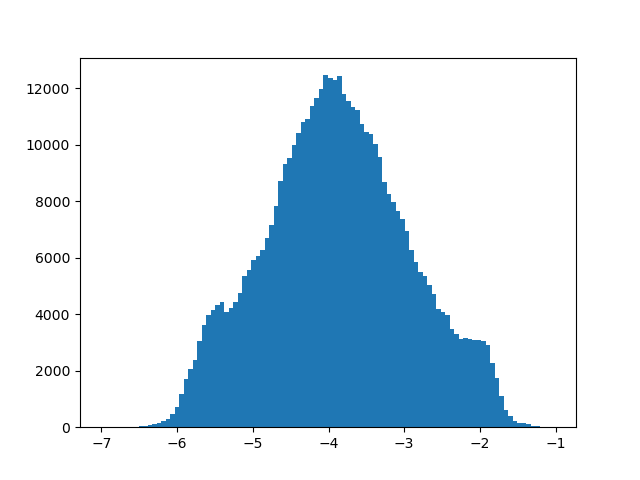} \caption{Feature 3} \end{subfigure}
	
	\begin{subfigure}[t]{0.30\textwidth} \includegraphics[width=\textwidth]{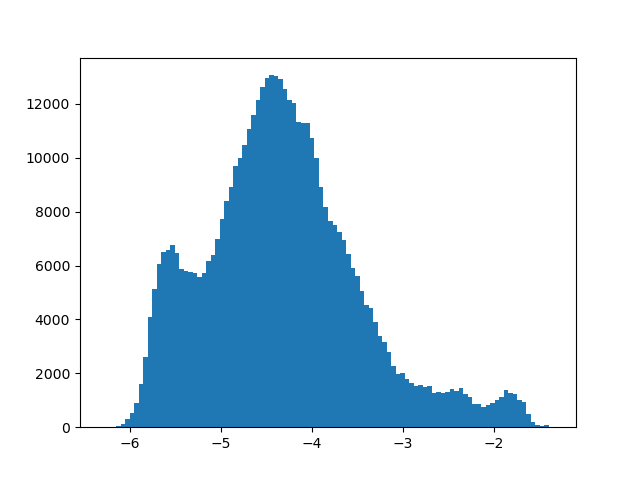} \caption{Feature 4} \end{subfigure}
	\begin{subfigure}[t]{0.30\textwidth} \includegraphics[width=\textwidth]{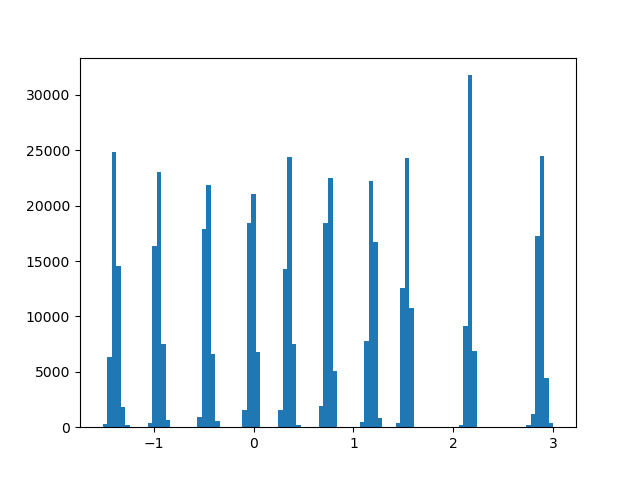} \caption{Feature 5 (Floor Hue)} \end{subfigure} 
	\begin{subfigure}[t]{0.30\textwidth} \includegraphics[width=\textwidth]{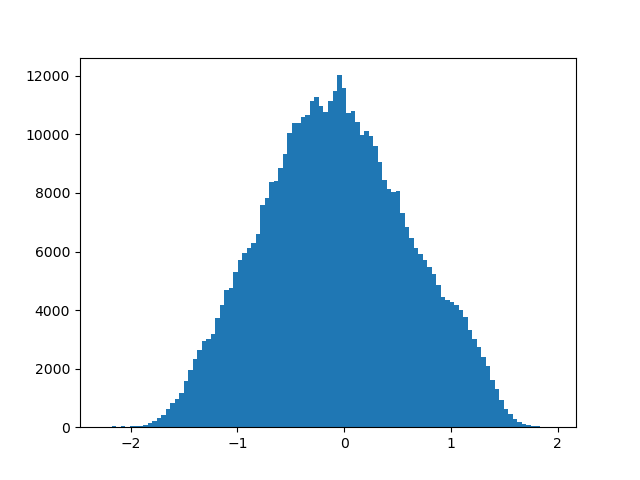} \caption{Feature 6} \end{subfigure} 
	
	\caption{Histograms of each latent feature for best performing TC-VCAE for disentanglement on 3D Shapes.}
	\label{fig:tcvcae_histograms}
\end{figure*}

\begin{figure*}
	\centering
	\begin{subfigure}[t]{0.30\textwidth} \includegraphics[width=\textwidth]{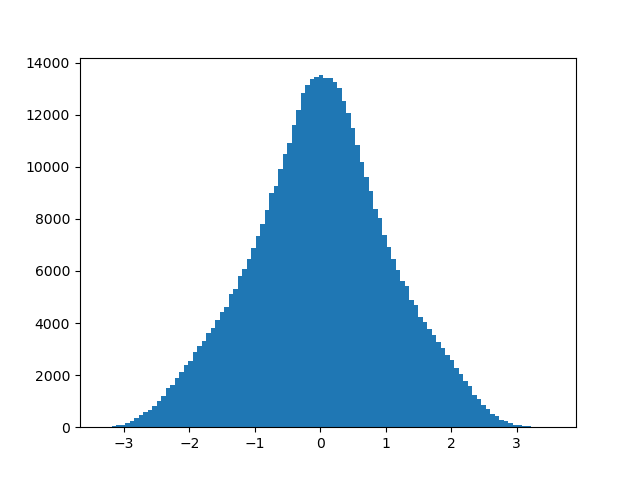} \caption{Feature 1} \end{subfigure}
	\begin{subfigure}[t]{0.30\textwidth} \includegraphics[width=\textwidth]{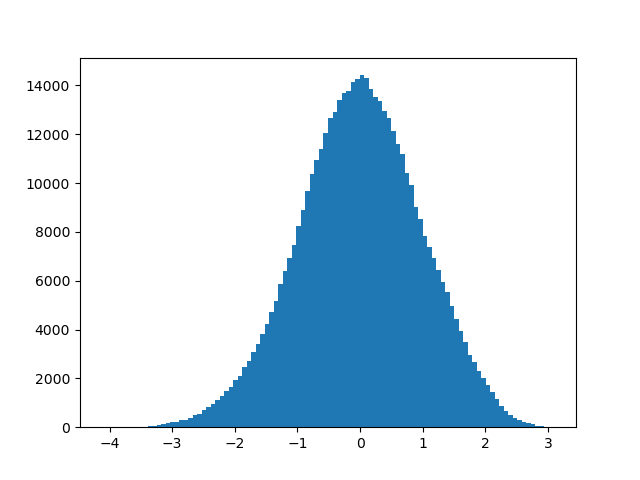} \caption{Feature 2} \end{subfigure}
	\begin{subfigure}[t]{0.30\textwidth} \includegraphics[width=\textwidth]{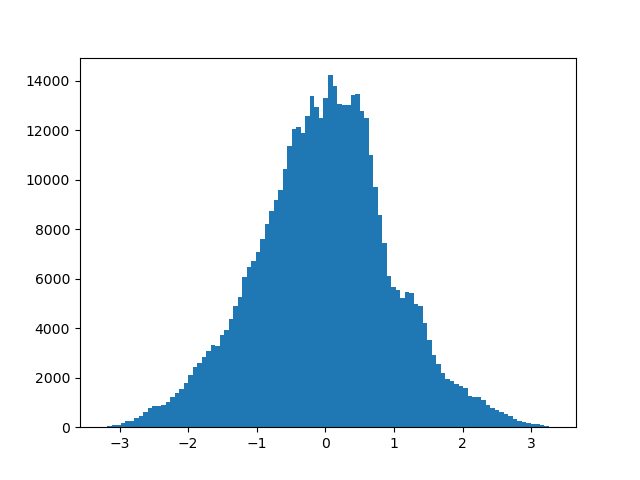} \caption{Feature 3} \end{subfigure}
	
	\begin{subfigure}[t]{0.30\textwidth} \includegraphics[width=\textwidth]{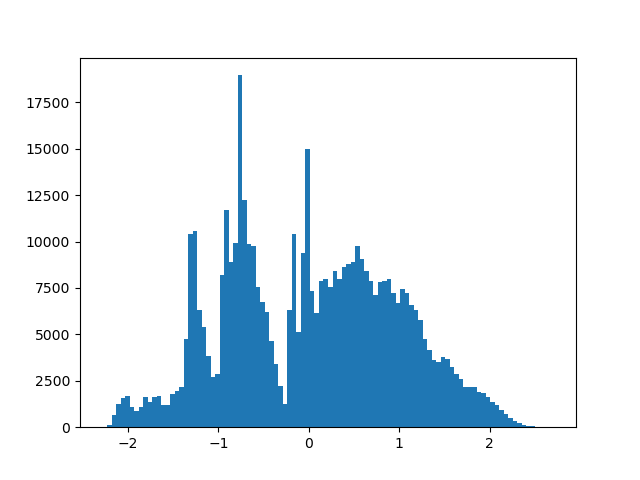} \caption{Feature 4} \end{subfigure}
	\begin{subfigure}[t]{0.30\textwidth} \includegraphics[width=\textwidth]{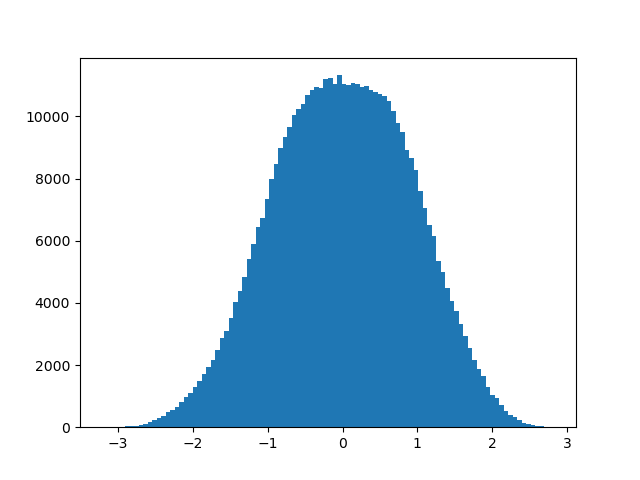} \caption{Feature 5} \end{subfigure} 
	\begin{subfigure}[t]{0.30\textwidth} \includegraphics[width=\textwidth]{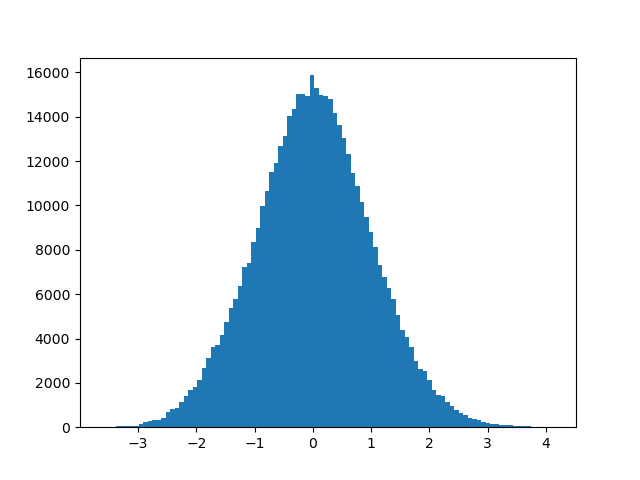} \caption{Feature 6} \end{subfigure} 
	
	\caption{Histograms of each latent feature for best performing TC-WAE for disentanglement on 3D Shapes.}
	\label{fig:tcwae_histograms}
\end{figure*}

\begin{figure*}
	\centering
	\begin{subfigure}[t]{0.30\textwidth} \includegraphics[width=\textwidth]{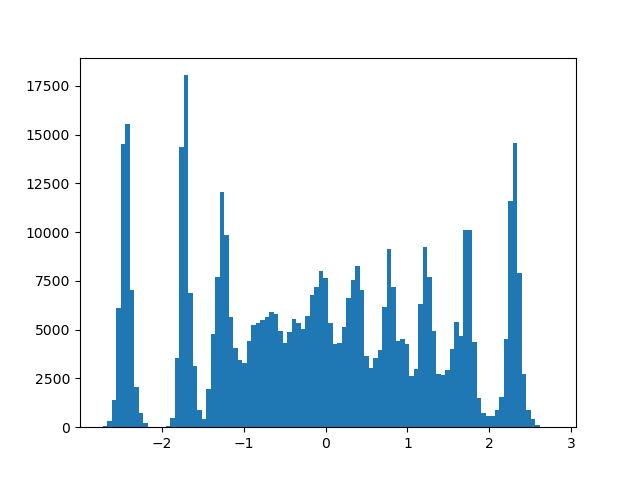} \caption{Feature 1} \end{subfigure}
	\begin{subfigure}[t]{0.30\textwidth} \includegraphics[width=\textwidth]{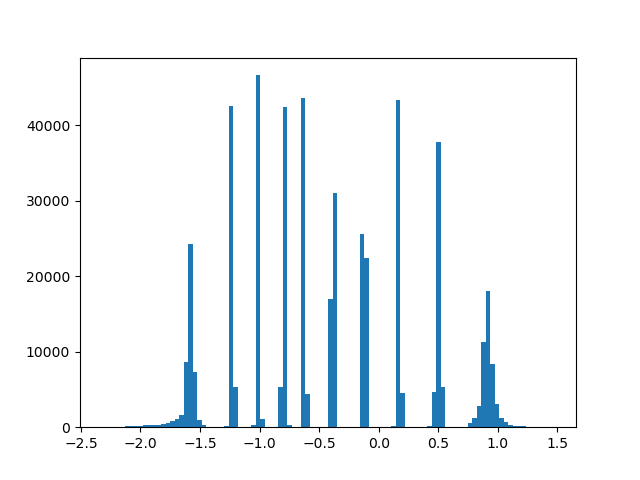} \caption{Feature 2 (Floor Hue)} \end{subfigure}
	\begin{subfigure}[t]{0.30\textwidth} \includegraphics[width=\textwidth]{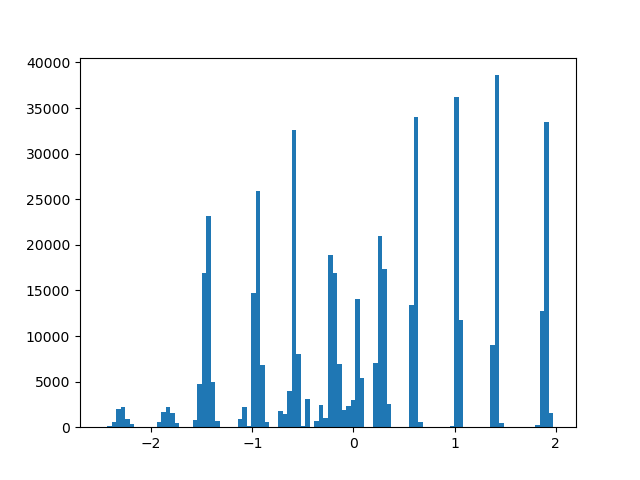} \caption{Feature 3 (Floor Hue)} \end{subfigure}
	
	\begin{subfigure}[t]{0.30\textwidth} \includegraphics[width=\textwidth]{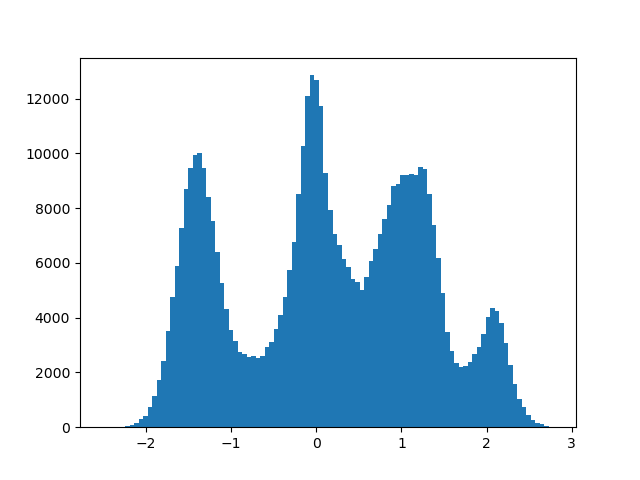} \caption{Feature 4} \end{subfigure}
	\begin{subfigure}[t]{0.30\textwidth} \includegraphics[width=\textwidth]{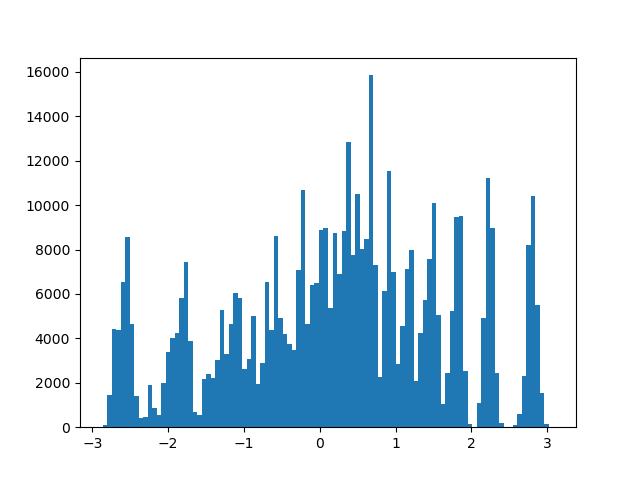} \caption{Feature 5 (Orientation)} \end{subfigure} 
	\begin{subfigure}[t]{0.30\textwidth} \includegraphics[width=\textwidth]{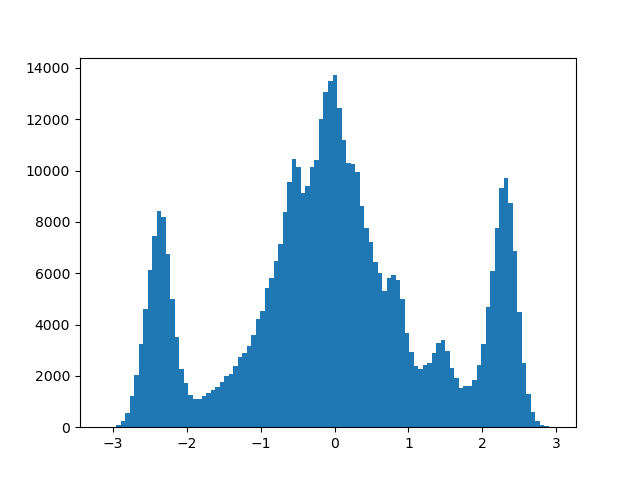} \caption{Feature 6} \end{subfigure} 
	
	\caption{Histograms of each latent feature for best performing FactorVAE for disentanglement on 3D Shapes.}
	\label{fig:factorvae_histograms}
\end{figure*}

In this section, we display further results from the disentanglement experiments presented in section \ref{sec:vcae_disentanglemetn}. Figure \ref{fig:tcvcae_histograms} shows a histogram of each latent feature learned by TC-VCAE, of which we first focus on Features 1, 2 and 5 which correspond to the room orientation, wall hue and floor hue respectively. This correspondence can be seen by inspecting figure \ref{fig:long_traversals_tcvcae}. The histograms show that features 1,2 and 5 have distributions with 15, 10 and 10 modes respectively. The number of modes for each distribution corresponds to the number of settings of that parameter when generating the data set. For example, there are 15 different settings of the room orientation in the dataset, which corresponds to the 15 modes in feature 1. Similarly, for wall and floor hue, there are ten possible settings in the dataset, again represented by ten modes in features 2 and 5 respectively. Careful inspection of the histograms and traversal in figures \ref{fig:factorvae_histograms} \& \ref{fig:long_traversals_factorvae} respectively, reveals that the same holds true for FactorVAE. However, the same does not hold true for TC-WAE, as seen by the histograms and traversal in figures \ref{fig:tcwae_histograms} \& \ref{fig:long_traversals_tcwae}  

Figure \ref{fig:long_traversals} shows three enlarged traversals, one for the best performing TC-VCAE, TC-WAE and FactorVAE. These results demonstrate that both TC-VCAE and FactorVAE can generalise to settings of orientation, wall hue and floor hue that were not present in the dataset as in this case, we have 30 settings.

\end{document}